\documentclass[sn-apa]{sn-jnl}% APA Reference Style
\usepackage{tabularx}
\usepackage{subcaption}
\AtBeginDocument{\renewcommand{\APACrefYearMonthDay}[3]{\BBOP{#1}\BBCP}}

\jyear{2022}%

\theoremstyle{thmstyleone}%
\begin{document}

\title[Disruption Detection for a CDSCT Using Hybrid Deep Learning]{Disruption Detection for a Cognitive Digital Supply Chain Twin Using Hybrid Deep Learning}


\author*[1,2]{\fnm{Mahmoud} \sur{Ashraf}}\email{mahmoud.ashraf@ejust.edu.eg}
%\orcidlink{0000-0002-5617-3213}

\author[1,2]{\fnm{Amr} \sur{Eltawil}}\email{eltawil@ejust.edu.eg}
%\orcidlink{0000-0001-6073-8240}

\author[1,2]{\fnm{Islam} \sur{Ali}}\email{islam.ali@ejust.edu.eg}
%\orcidlink{0000-0002-7344-6113}

\affil*[1]{\orgdiv{Department of Industrial Engineering and Manufacturing}, \orgname{Egypt-Japan University of Science and Technology}, \orgaddress{\city{New Borg El-Arab}, \state{Alexandria}, \country{Egypt}}}

\affil[2]{\orgdiv{Department of Production Engineering}, \orgname{Alexandria University}, \orgaddress{\state{Alexandria}, \country{Egypt}}}


\abstract{\textbf{Purpose:} Recent disruptive events, such as COVID-19 and Russia-Ukraine conflict, had a significant impact of global supply chains. Digital supply chain twins have been proposed in order to provide decision makers with an effective and efficient tool to mitigate disruption impact. \\ \textbf{Methods:} This paper introduces a hybrid deep learning approach for disruption detection within a cognitive digital supply chain twin framework to enhance supply chain resilience. The proposed disruption detection module utilises a deep autoencoder neural network combined with a one-class support vector machine algorithm. In addition, long-short term memory neural network models are developed to identify the disrupted echelon and predict time-to-recovery from the disruption effect. \\ \textbf{Results:} The obtained information from the proposed approach will help decision-makers and supply chain practitioners make appropriate decisions aiming at minimizing negative impact of disruptive events based on real-time disruption detection data. The results demonstrate the trade-off between disruption detection model sensitivity, encountered delay in disruption detection, and false alarms. This approach has seldom been used in recent literature addressing this issue.}




\maketitle

\section{Introduction}
\label{sec:introduction}

Local and global crises severely impact global supply chains. Hurricane Katrina in 2006, the Japanese tsunami in 2011, COVID-19 in late 2019, and the Suez Canal blockage in 2021 disrupted the flow of goods and materials in global supply chains. Recent power outages and industrial shutdowns in China have affected many supply chains with limited supply and long delays \citep{b24}. Furthermore, climate change risks may evolve and disrupt global supply chains through natural disasters, resulting in plant shutdowns and disruptions to mining operations and logistics \citep{b26}. Finally, the Russia-Ukraine conflict is expected to adversely impact many supply chains worldwide and global logistics \citep{b38}.

In 2021, 68\% of supply chain executives reported constantly facing disruptive events since 2019 \citep{b28}. Therefore, proper disruption management is vital to minimise negative disruption impacts and avoid supply chain collapse. Supply chain disruption management refers to the approaches and policies adopted to recover from unexpected disruptive events which cause a high adverse impact on supply chain performance and are characterised by low occurrence frequency \citep{b3}. Some disruptive events, such as supplier unavailability, can have a prolonged impact during the post-disruption period due to delayed orders and backlogs. Supply Chain Resilience (SCR) refers to the supply chain’s ability to withstand, adapt, and recover from disruptions to fulfil customer demand and maintain target performance \citep{b2}. For dynamic systems, SCR is a performance-controlled systemic property and goal-directed. In other words, disruption absorption allows for maintaining the intended performance in the event of a disruption. At the same time, the feedback control embodied in recovery control policies makes SCR self-adaptable \citep{b3}.

SCR considers disturbances in the supply chain, such as supplier unavailability and disruption impact on supply chain performance. Moreover, SCR seeks to restore normal operations by adopting recovery policies. As a result, SCR guarantees the firm’s survival after severe adverse events. Resilience may be realised by (1) redundancies, such as subcontracting capabilities and risk mitigation stocks, (2) recovery flexibility to restore regular performance, and (3) end-to-end supply chain visibility \citep{b3}.

With the evolution of Industry 4.0, many businesses were encouraged to carry out the transition towards digitalisation. \citet{b1} predicted that by 2023, at least half of the world’s largest corporations would be employing Artificial Intelligence (AI), advanced analytics, and the Internet of Things (IoT) in supply chain operations. Big Data Analytics (BDA) advancements and real-time data availability offered by IoT technologies resulted in the emergence of Digital Twins (DTs). A DT is a digital representation of a real-world physical system \citep{b17}.

A Digital Supply Chain Twin (DSCT), as defined by \citet{b10}, is “a computerised model of the physical system representing the network state for any given moment in real-time”. The DSCT imitates the supply chain, including any vulnerability, in real-time. This real-time representation helps improve SCR through an extensive end-to-end supply chain visibility based upon logistics, inventory, capacity, and demand data \citep{b11}.

DSCTs can improve SCR, minimise risks, optimise operations, and boost performance \citep{b14}. DTs provide up-to-date real-time data which reflects the most recent supply chain state. Real-time data allows for the early detection of supply chain disruptions and rapid response through recovery plans. Moreover, optimisation engines integration with DTs enable making the most cost-effective operational decisions \citep{b21}.

The concept of Cognitive Digital Twins (CDTs) has emerged during the past few years which refers to the DTs that possess additional capabilities, such as communication, analytics, and cognition \citep{b29}. CDTs have been firstly introduced in the industry sector in 2016, followed by several attempts to provide a formal definition of CDTs \citep{b29}. For instance, \citet{b49} defined CDTs as “DTs with augmented semantic capabilities for identifying the dynamics of virtual model evolution, promoting the understanding of inter-relationships between virtual models and enhancing the decision-making”. CDTs which utilise machine learning can sense and detect complex and unpredictable behaviours. Therefore, a Cognitive Digital Supply Chain Twin (CDSCT) permits disruption detection in the supply chain and quick deployment of recovery plans in real-time upon disruption detection.

Motivated by recent global supply chain disruptions, digital transformation efforts, and absence of operational frameworks that utilize CDSCT for disruption detection and time-to-recovery prediction from the literature, this paper introduces a framework to help enhance Supply Chain Resilience (SCR) through decision support by adopting Digital Supply Chain Twins (DSCTs), building upon the introduced conceptual framework by \citet{b11} Additionally, the adoption of data-driven AI models in DSCTs enable monitoring supply chain state that help detect supply chain disruptions in real-time and optimising recovery policies to recover from these disruptions. Real-time disruption detection enables the decision-makers to respond quickly to disruptions through early and efficient deployment of recovery policies. AI models play an important role in discovering abnormal patterns in data. As a result, this paper introduces a hybrid deep learning approach for disruption detection in a make-to-order three-echelon supply chain. The proposed approach is presented within a CDSCT framework to improve SCR through real-time disruption detection. The introduced approach allows the decision-makers to identify the disrupted echelon and obtain an estimate of the Time-To-Recovery (TTR) from a disruptive event upon disruption detection.

The remainder of this paper is organised as follows. Section~\ref{sec:literature} reviews the relevant literature. Then, section~\ref{sec:problem} introduces and describes the problem at hand. Afterwards, section~\ref{sec:background} demonstrates pertinent machine learning concepts, followed by section~\ref{sec:methodology}, demonstrating the development steps. The results are shown in section~\ref{sec:results} followed by section~\ref{sec:practical implications}, demonstrating the managerial implications. Finally, section~\ref{sec:conclusion} provides concluding remarks, current research limitations, and directions for future work.

\section{Review of literature}
\label{sec:literature}

\subsection{Supply chain resilience}

Many scholars proposed several signal-based approaches to evaluate SCR \citep{b40,b44,b43,b39,b45}. The proposed approaches involved simple models, such as simple aggregation models, and sophisticated models, such as deep learning. An aggregation-based approach was introduced to evaluate operational SCR \citep{b4}. A single evaluation metric across multiple tiers in a multi-echelon supply chain was developed by aggregating several transient response measures. The transient response represents the change in supply chain performance due to a disruptive event. The transient response measures evaluated supply chain performance across multiple dimensions. These dimensions were (1) TTR, (2) disruption impact on performance, (3) performance loss due to disruption, and (4) a weighted-sum metric to capture the speed and shape of the transient response. This approach could explain the performance response to supply chain disruptions better than individual dimensions of resilience at the single-firm level.

A system dynamics-based approach was proposed to quantify SCR at a grocery retailer \citep{b5}. SCR was evaluated based on the supply chain response to the dynamic behaviour of stock and shipment in a distribution centre replenishment system. Considering the inherent non-linear system behaviour eliminates preliminary analysis of non-linearity effects which helps simulate complex supply chains \citep{b6}.

A hierarchical Markov model was introduced to integrate advance supply signals with procurement and selling decisions \citep{b7}. The proposed model captured essential features of advance supply signals for dynamic risk management. In addition, the model could be used to make a signal-based dynamic forecast. The strategic relationship between signal-based forecast, multi-sourcing, and discretionary selling was revealed. However, future supply volatility and variability are expected to affect the future supply forecast. The findings revealed a counter-intuitive insight. A model that disregards both volatility and variability of the uncertain future supply might outperform the one that considers the variability of the uncertain future supply. Finally, a signal-based dynamic supply forecast was recommended under considerable supply uncertainty and a moderate supply-demand ratio.

Deep learning models for enhancing SCR could outperform the classical models. A deep learning approach was introduced based on Artificial Neural Networks (ANNs) \citep{b8}. This approach aims at identifying disruptions related to temperature anomalies in the cold supply chain during transport. The ANN-based model was compared to another approach based on BDA and mathematical modelling. Based on a simulation model and a real-world case, the ANN-based model outperformed the other model based on BDA and mathematical modelling.

Moreover, hybrid deep learning models could outperform deep learning models for anomaly detection. A hybrid-deep learning approach was presented to detect anomalies in a fashion retail supply chain \citep{b9}. The hybrid deep learning model involved a deep Long-Short term memory (LSTM) autoencoder and classic machine learning to extract meaningful information from the data. Then, semi-supervised machine learning was applied in the shape of a One-Class Support Vector Machine (OCSVM) algorithm to detect sales anomalies. Based on a real case for a company in France, the results showed that hybrid approaches could perform better than deep learning-based approaches.

\subsection{Digital supply chain twins for enhancing supply chain resilience}

Several studies extended the application of DSCTs in many aspects to support decision-making and enhance SCR. A machine learning approach was introduced to improve SCR through resilient supplier selection in a DT-enabled supply chain \citep{b12}. The introduced approach could analyse the supplier performance risk profiles under uncertainty through data-driven simulation for a virtual two-echelon supply chain. The results revealed that combining machine learning-based methods with DSCT could enhance SCR, especially when redesigning the supply network.

A notion of DSCT to support decision-making and improve SCR was explained in \citep{b10}. The interrelationships between supply chain digital technology and disruption risk effects in the supply chain were investigated. Then, a framework for risk management in supply chain management was introduced. The results indicated that future decision support systems would utilise DSCTs and digital technologies, such as IoT and BDA. As a result, the available real-time data could provide information regarding the scope and impact of disruptions. The feedback from DSCTs could be used to restore the pre-disruption performance by testing different policies. The integration between BDA and a DT for an automotive supply chain was introduced to support decision-making and adapt to new scenarios in real-time \citep{b13}.

Another framework based on real-time disruption detection was presented to support decision-making for a DSCT for disruption risk management \citep{b11}. This framework would enable efficient deployment of recovery policies, reliable disruption scenarios creation for supply chain risk analysis, and revealing the connections between risk data, disruption modelling, and performance evaluation.

The weaknesses in SCR modelling were highlighted in the face of foreseeable disruptions \citep{b16}. The findings showed that DSCTs could better allow decision-makers to evaluate efficiency/resilience trade-offs. Furthermore, during the post-disruption phase, DTs can help optimise system performance.

Corresponding to the COVID-19 impact on global supply chains, DSCTs were used to examine the effect of a real-life pandemic disruption scenario on SCR for a food retail supply chain \citep{b15}. The results uncovered the underlying factors that affect supply chain performance, such as pandemic intensity and customer behaviour. The findings assured the importance of DSCTs for building resilient supply chains.

\subsection{Cognitive digital twins}

Many scholars introduced different architectures and implementations for CDTs in various fields, such as condition monitoring of assets, real-time monitoring of finished products for operational efficiency, and supporting demand forecasting and production planning \citep{b48}. In the field of manufacturing and supply chains, introduced architectures focused on detecting anomalous behaviour in manufacturing systems, improving operations, and minimizing cost across the supply chain \citep{b17, b18}. A CDT architecture was proposed for real-time monitoring and evaluation for a manufacturing flow-shop system \citep{b17}. The CDT platform could forecast and identify abnormalities using the available data from interconnected cyber and physical spaces. In addition, another architecture was introduced for a shop floor transportation system to predict and identify anomalous pallet transportation times between workstations \citep{b18}. Based on two different showcases, both architectures showed that CDTs could improve operations through optimal scheduling in real-time and enhanced resource allocation.

A CDT framework was introduced for logistics in a modular construction supply chain \citep{b19}. The proposed CDT could predict logistics-related risks and arrival times to reduce costs using IoT and Building Information Modeling (BIM). Furthermore, an approach for a CDT was proposed in agile and resilient supply chains \citep{b20}. The CDT could predict trends in dynamic environments to guarantee optimal operational performance. This approach was elaborated through a connected and agile supply chain. The deployed model considers collaboration among different actors as enablers for information exchange, processing, and actuation.

In addition, a deep learning-based approach has been introduced to predict TTR in a three-echelon supply chain \citep{b47}. The introduced approach was presented within a theoretically proposed CDSCT framework to enhance SCR. Obtained results showed that predicted TTR values tend to be relatively lower than the actual values at early disruption stages, then improve throughout the progression of the disruption effect on the supply chain network.

It has been observed from the literature that many recent contributions were directed towards SCR in response to the COVID-19 pandemic impact on global supply chains. Many scholars were concerned with quantifying SCR and deploying DSCTs frameworks. On the one hand, deep learning-based models outperformed the classic ones for enhancing SCR. On the other hand, few contributions concerned with enhancing SCR through deep learning-based techniques in a CDSCT environment have been observed. In addition, the literature emphasized the role of CDSCTs in the field of supply chain disruption management. However, few contributions on the implementation of different CDSCT modules for disruption detection was observed. Therefore, this paper contributes to the literature through developing the CDSCT enabling modules for disruption detection.

This paper extends the proposed framework by \citet{b47} through incorporating an additional layer for disrupted echelon identification. Furthermore, this paper extends their work by introducing: (1) a hybrid deep learning approach for disruption detection and (2) deep learning-based model for disrupted echelon identification. The introduced approaches are presented as sub-modules of CDSCT for a make-to-order virtual supply chain. In addition, this paper reconsiders inputs for the TTR prediction modules with the aim of obtaining better TTR estimates.

This study tries to answer two research questions. The main research question is “Is there a way to exploit the benefit of cognitive digital twins in the field of supply chain disruption management?” The second research question is “How to validate the introduced framework for incorporating cognitive digital twins into supply chain disruption management. The first research question is addressed by introducing a CDSCT framework that allows early disruption detection in a CDT-enabled make-to-order virtual supply chain. Early disruption detection is enabled through a hybrid deep learning-based approach using a deep autoencoder neural network and the OCSVM algorithm. In addition to early disruption detection, the CDSCT permits disrupted echelon identification and TTR prediction. The first research question is addressed throughout the introduced framework, while the second research question is addressed throughout system the implementation.

\section{Problem statement}
\label{sec:problem}

\begin{figure}
\begin{center}
	\begin{subfigure}{\linewidth}
		\includegraphics[width=\linewidth]{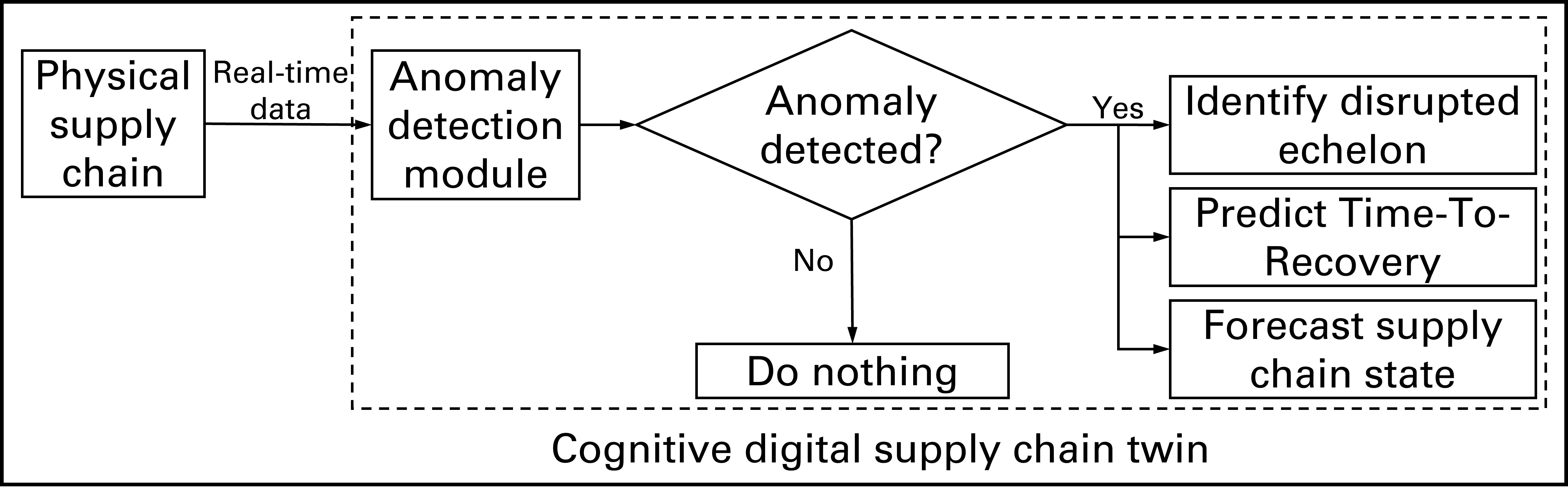}
		\caption{During the operational phase.}
		\label{fig:1a}
	\end{subfigure}
	\begin{subfigure}{\linewidth}
		\vspace{2mm}
		\includegraphics[width=\linewidth]{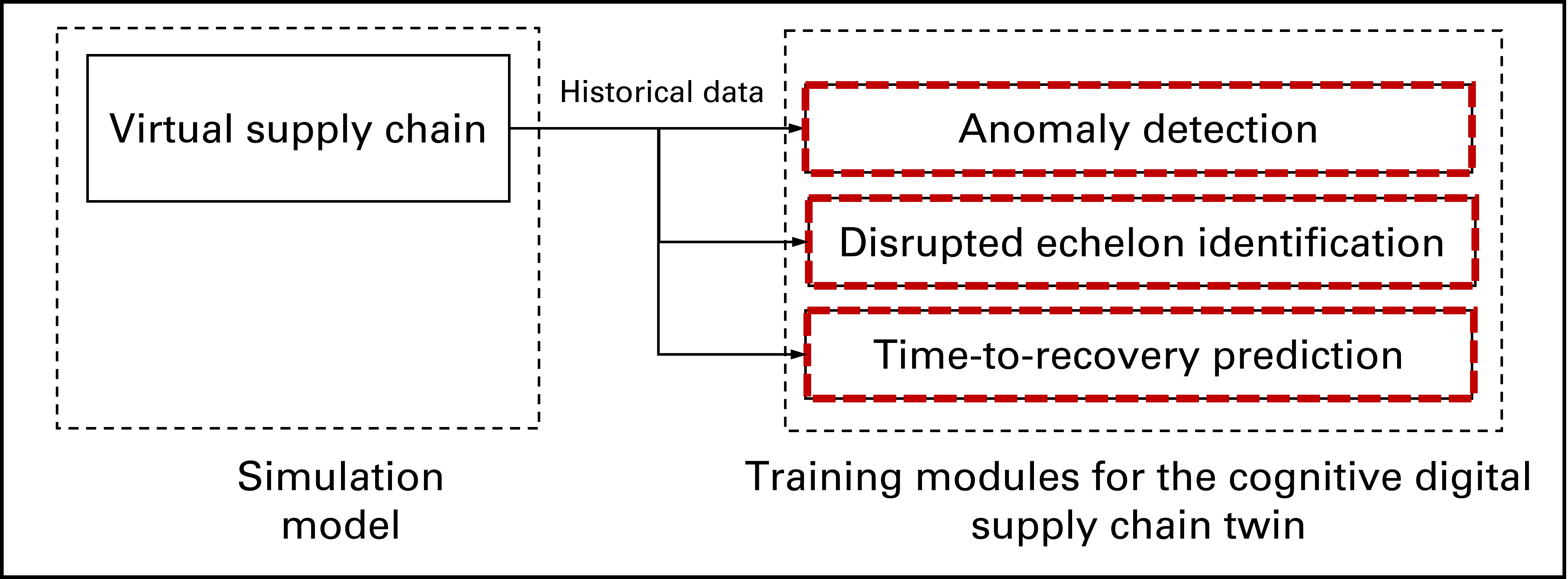}
		\caption{During the training phase using a virtual supply chain.}
		\label{fig:1b}
	\end{subfigure}
	\caption{The cognitive digital supply chain twin framework.}
	\label{fig:1}
\end{center}
\end{figure}

This paper introduces a hybrid deep learning approach for disruption detection within a CDSCT framework to enhance SCR. This approach involves (1) a training phase and (2) an operational phase. The \emph{training phase} involves training the disruption detection module and models for disrupted echelon identification and TTR prediction. After the training phase, the CDSCT can detect supply chain disruptions, identify disrupted echelons, and predict TTR from disruptions. Figure~\ref{fig:1a} demonstrates the CDSCT during the \emph{operational phase}. Supply chain disruptions are detected based on a real-time data stream from an existing supply chain. The literature indicated that real-time data, enabled by IoT, is collected in multiple means, such as sensors and RFID tags \citep{b11}. Then, the disrupted echelon is identified upon disruption detection, and TTR estimates are obtained. In addition, future supply chain states can be forecasted due to the disruption impact.

Needed supply chain data for training the anomaly (disruption) detection module and TTR prediction model can be obtained from multiple sources. These sources include historical records, real-time data from an IoT-enabled supply chain, or a simulation model depicting a real or a virtual system. Figure~\ref{fig:1b} demonstrates the framework during the training phase. This phase involves the training based on historical data feed representing the supply chain performance in normal and disrupted states. The disrupted echelon is identified upon disruption detection. Then, a TTR estimate is obtained after feeding the labelled training data to the CDT. In practice, sufficient historical records of disruptions for training purpose may be unavailable due to the unpredictability and low occurrence frequency of disruptive events. In such cases, simulation modelling becomes the most convenient tool for augmenting the training data required for the development of machine learning models. This paper uses simulation modelling to simulate different disruption scenarios. In addition, the developed simulation model is used to generate the required data for training the disruption detection module for a make-to-order virtual three-echelon supply chain.

\section{Methodology}
\label{sec:background}

\subsection{Deep autoencoders}

\begin{figure}
	\begin{center}
		\includegraphics[width=.8\linewidth]{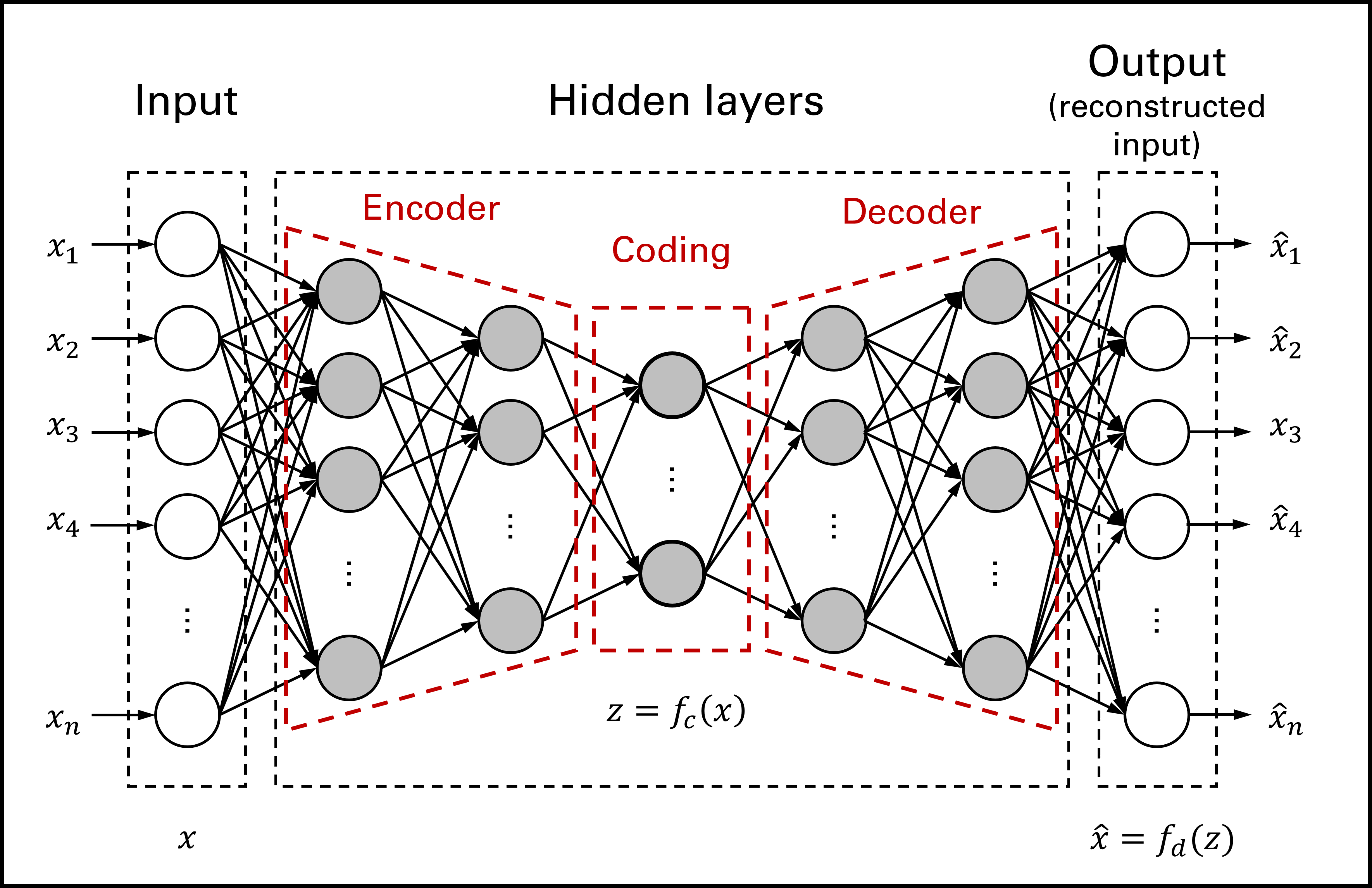}
		\caption{Autoencoder architecture.}
		\label{fig:5}
	\end{center}
\end{figure}

An autoencoder is a special type of feedforward neural network trained to copy its input to its output by representing its input as coding \citep{b30}. Autoencoder consists of three main components, other than the input and output, (1) encoder, (2) coding, and (3) decoder, Figure~\ref{fig:5}. Input data compression and decompression through the encoder and decoder, respectively, makes autoencoders ideal for applications involving dimensionality reduction and feature extraction. The coding, $z$, represents the compressed representation of the input vector x, which contains the most representative information of the input.

An autoencoder with three or more hidden layers in the encoder or the decoder network is considered a deep one \citep{b31}. The autoencoder is trained to minimise the reconstruction error between the input $x$ and the output $\hat{x}$. It is expected that an autoencoder trained on normal (non-disrupted) data will result in a high reconstruction error when given anomalous (disrupted) data \citep{b32}. Therefore, an autoencoder neural network is used for the problem at hand of disruption detection.

\subsection{The one-class support vector machine algorithm}

OCSVM is a machine learning algorithm used for binary classification and anomaly detection. Anomaly detection refers to discovering outliers or abnormalities embedded in a large amount of normal data \citep{b33}. OCSVM works in a semi-supervised manner when considering anomaly detection as the application area. During training an OCSVM, it learns to construct the boundary that separates observations under normal conditions from abnormal observation. The work proposed by \citet{b34,b33} introduced the inherent mechanism of OCSVM for anomaly detection in a more detailed manner. OCSVM is usually trained using normal points representing the positive class because full consideration of all disruption scenarios is quite impossible. Then, during operation, the OCSVM checks whether new data points belong to the normal class or not. Suppose an input data point is considered anomalous. In that case, it lies outside the boundary and belongs to the other class, usually referred to as the negative class (anomalous).

The OCSVM algorithm is applied for automatic disruption (anomaly) detection in a three-echelon supply chain. As a binary classification and anomaly detection algorithm, OCSVM was chosen as it enables disruption detection without a prohibitively extensive study of all potential disruption scenarios. The first principal component of the reconstruction error obtained from the autoencoder is used as the input to the OCSVM algorithm. The OCSVM algorithm eliminates the need for statistical analyses to set a threshold above which a data point is considered anomalous. In addition, the OCSVM algorithm does not necessitate any specific assumptions about the data, i.e., reconstruction error is normally distributed \citep{b9}.

\subsection{Long-short term memory neural networks}

\begin{figure}
	\begin{center}
		\begin{subfigure}{0.3\linewidth}
			\includegraphics[width=\linewidth]{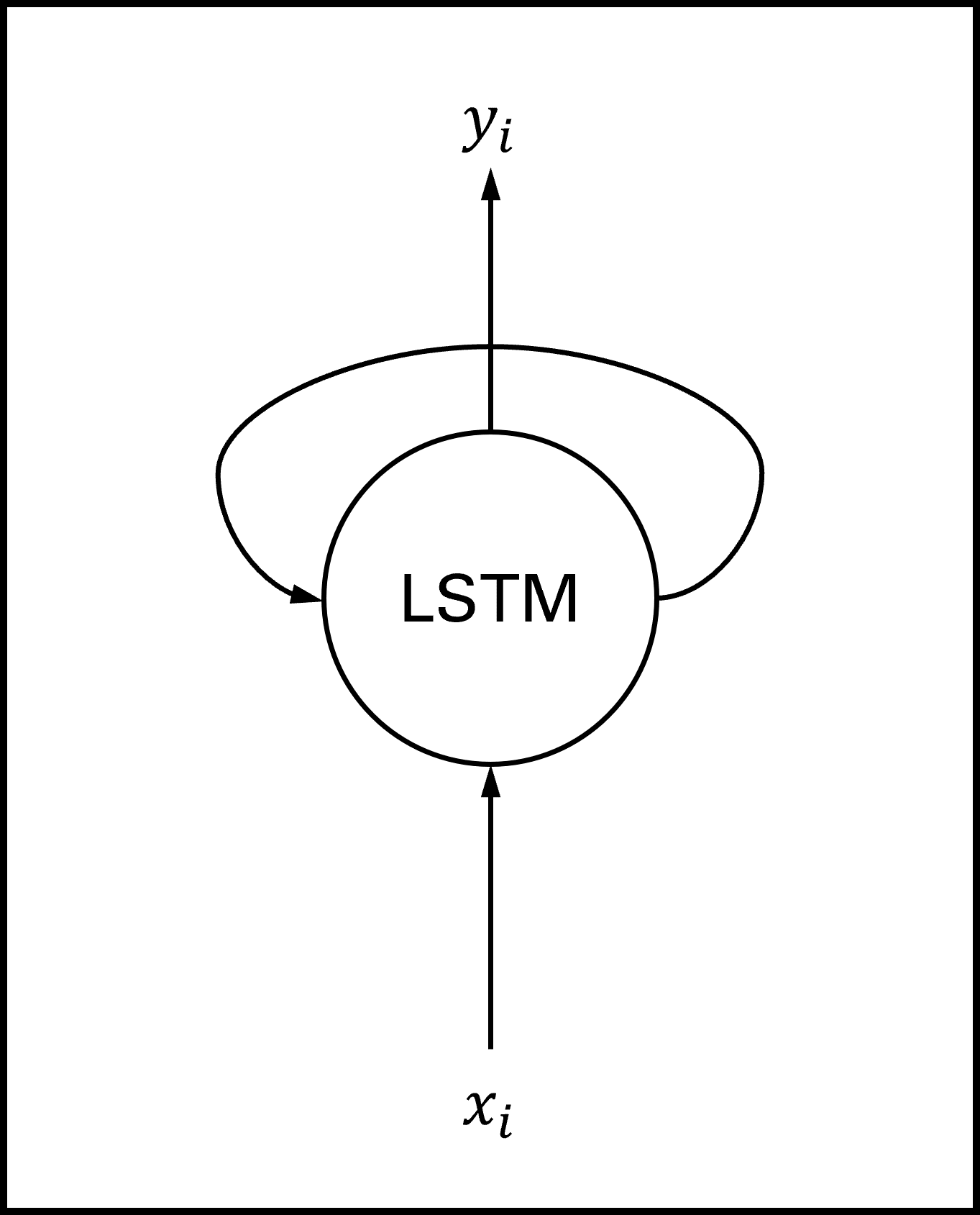}
			\caption{Compressed single neuron.}
			\label{fig:6a}
		\end{subfigure}
		\begin{subfigure}{0.59\linewidth}
			\includegraphics[width=\linewidth]{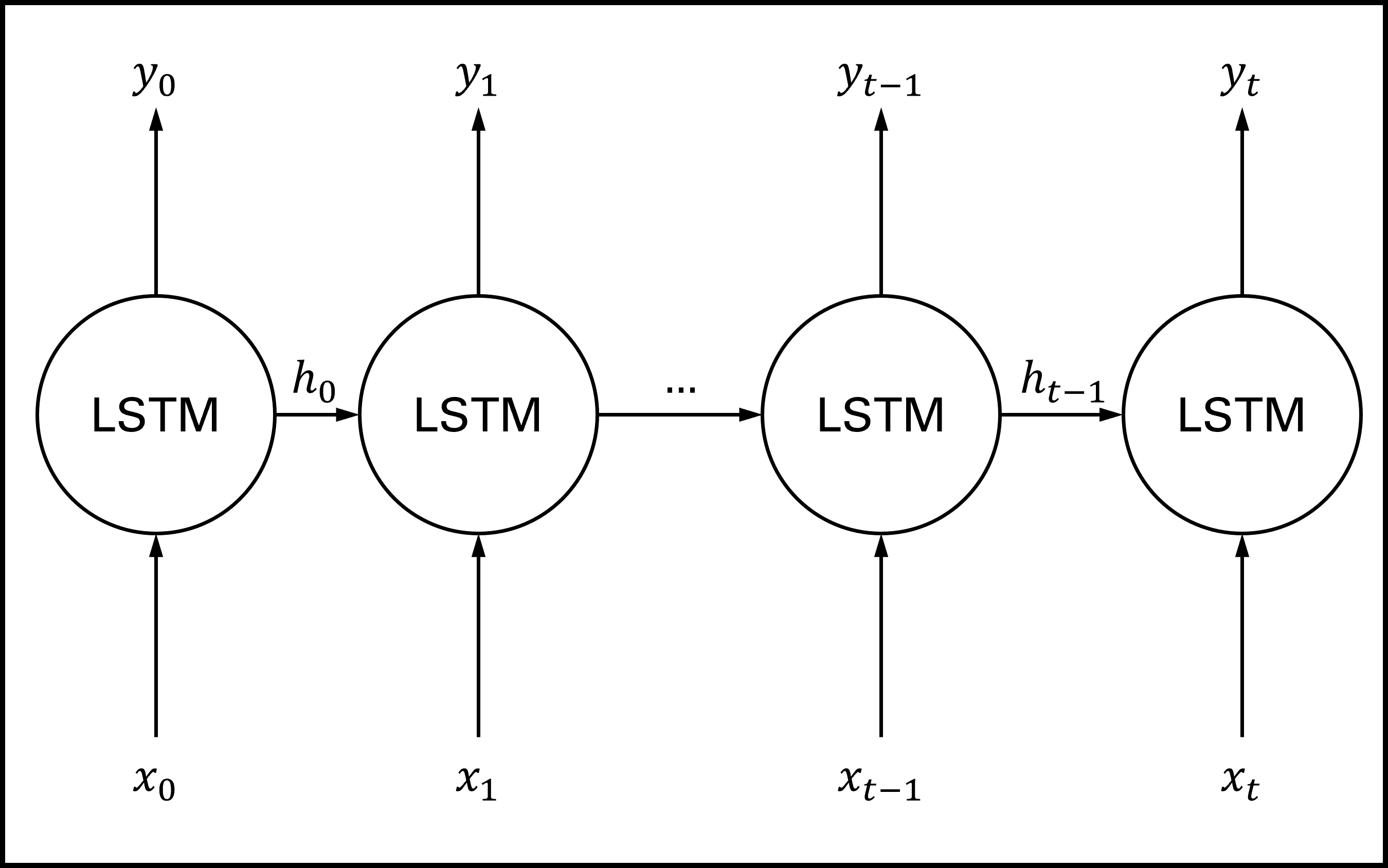}
			\caption{A single neuron unfolded through time.\newline}
			\label{fig:6b}
		\end{subfigure}
		\caption{A long-short term memory neural network architecture.}
		\label{fig:6}
	\end{center}
\end{figure}

LSTM neural network is an important Recurrent Neural Networks (RNNs) class. LSTM neural networks were proposed by \citep{b36}. They provide memory to retain long-term dependencies among input data without suffering from the vanishing gradient problem \citep{b37}. Therefore, LSTM networks are suitable to represent sequential data, i.e., time series. A simple LSTM neural network of one neuron, Figure~\ref{fig:6a}, receives an input $x_{i}$, produces an output $y_{i}$, and resends that output to itself. When the LSTM neural network is unfolded through time, it has the form of a chain of repeated modules, Figure~\ref{fig:6b}. At each time step (frame) $i, i \in \left\{ 1, 2, ..., t \right\}$, that recurrent neuron receives the inputs $x_{i}$, and its output from the previous time step $h_{i-1}$, to produce an output $y_{i}$.

\section{System implementation}
\label{sec:methodology}

This section lays out the implementation steps for developing the proposed approach on a desktop computer with a 2.9 GHz Intel Core i7 processor and 8 GB RAM. A virtual supply chain is modelled as a discrete event simulation model using AnyLogic 8.7 simulation software. Machine learning models are developed using Python 3.8, Scikit-learn 0.24, and Keras 2.6. The training time for different models ranged between two and five hours.

\subsection{The virtual supply chain structure}

\begin{figure}
	\begin{center}
		\includegraphics[width=0.9\linewidth]{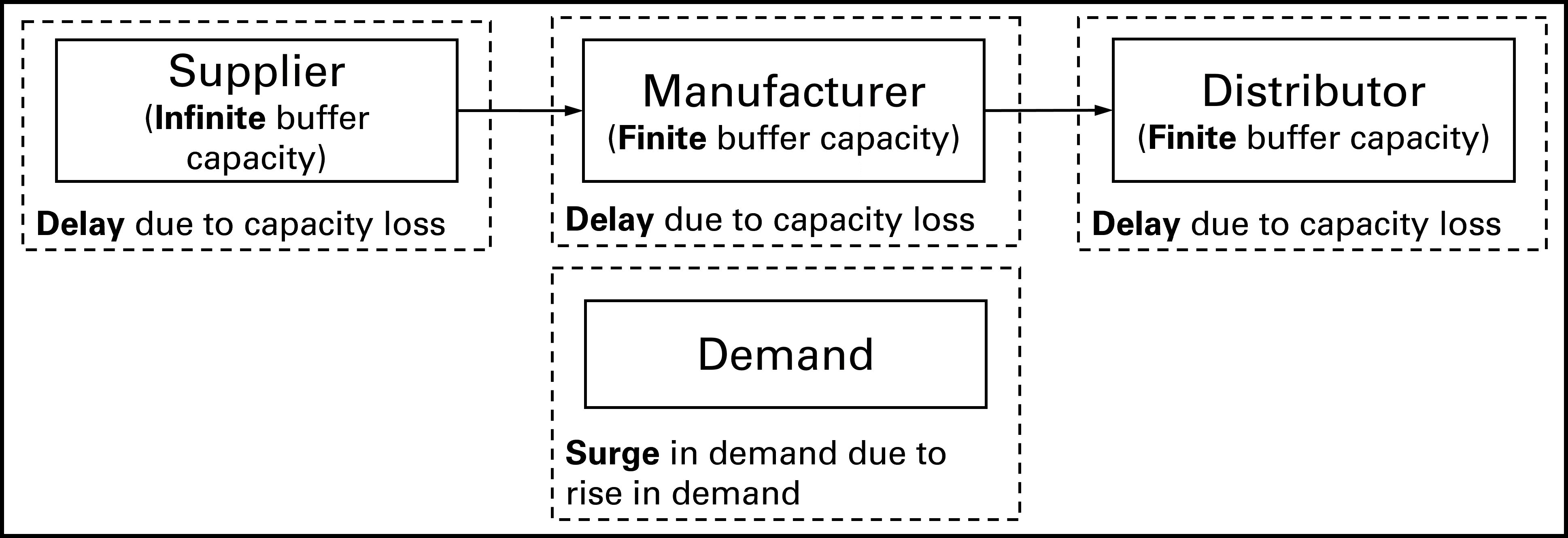}
		\caption{Virtual supply chain components with potential sources of disruptions.}
		\label{fig:3}
	\end{center}
\end{figure}

The three-stage flow line model with limited buffer capacity introduced by \citet{b27} is used to develop a make-to-order virtual three-echelon supply chain. It is assumed that there is a single product under consideration, and alternatives to any echelon are not available. Hence, the service protocol permits backlogging. Figure~\ref{fig:3} shows the main components of the virtual supply chain with potential sources of disruption.

A single supplier, manufacturer, and distributor constitute the three-echelon virtual supply chain. An additional component, \emph{demand}, corresponds to the initiated customer order quantity. After a customer order is generated, it enters a First-Come-First-Served (FCFS) queue waiting to be fulfilled. The customer order generation rate follows a Poisson distribution with a mean value of $\lambda$.

The \emph{supplier} provides the required raw material with a mean rate $\mu_{1}$. The supplier is assumed to have unlimited buffer capacity. In contrast, the remaining two echelons are assumed to have a limited buffer capacity of ten units. After the raw material is prepared and delivered to the \emph{manufacturer}, the products are manufactured with a processing rate $\mu_{2}$. Then, the customer order is ready to be fulfilled through the \emph{distributor} after being processed with a processing rate $\mu_{3}$. The processing rates at the supplier, manufacturer, and distributor are assumed to follow an exponential distribution.

Different scenarios are considered to account for the supply chain performance under normal and disrupted circumstances. The normal scenario is denoted by $S_{0}$, while potential disruption scenarios include unexpected failures at any single echelon $i$ and are denoted by $S_{i}, i \in \{ 1, 2, 3 \}$, where echelons 1, 2, and 3 correspond to the supplier, manufacturer, and distributor, respectively. In addition, the surge in demand scenario is considered and denoted by $S_{4}$. The simulation model parameters assumed values are shown in table~\ref{tab:1}.

\begin{table}
	\begin{center}
		\caption{Simulation model parameters.}
		\label{tab:1}
		\begin{tabularx}{0.8\linewidth}{lXlX}
			Parameter & Value \\
			\hline
			Number of replications, $N$ & 300 replications \\
			Replication length, $RL$ & 1095 days \\
			Warm-up period length, $WL$ & 180 days \\
			Arrival rate, Poisson($\lambda$) & 15 units/day \\
			Number of orders per arrival, $Q_{a}$ & 1 unit \\
			Supplier service rate, Poisson($\mu_{1})$ & 18 units/day \\
			Supplier buffer capacity, $q_{1}$ & $\infty$ \\
			Supplier server capacity, $c_{1}$ & 1 unit \\
			Manufacturer service rate, Poisson($\mu_{2}$) & 19 units/day \\
			Manufacturer buffer capacity, $q_{2}$ & 15 units \\
			Manufacturer server capacity, $c_{2}$ & 1 unit \\
			Distributor service rate, Poisson($\mu_{3}$) & 20 units/day \\
			Distributor buffer capacity, $q_{3}$ & 10 units \\
			Distributor server capacity, $c_{3}$ & 1 unit \\
			Disruption duration, $D_{d}$ & $D_{d} \in \left[ 30, 60 \right]$ days \\
			Disruption occurrence, $D_{t}$ & $D_{t} \in \left[300, 600\right]$ days \\
			Disrupted arrival rate, Poisson($\lambda_{d}$) & 30 units/day \\
			Disrupted processing rate, Poisson($\mu_{di}$) & $\mu_{di}$= 0 units/day \quad $\forall i \in \left\{ 1, 2, 3 \right\}$ \\
			\hline
		\end{tabularx}
	\end{center}
\end{table}

Several parameters and metrics reflecting the supply chain state and performance are monitored. The parameters include (1) the interarrival time, $T_{a}$, and (2) the processing time at echelon $i$, $T_{pi}, i \in \{ 1, 2, 3 \}$. Monitored metrics include (1) units in the system $WIP$, (2) queue length at echelon $i$, $L_{qi}, i \in \{ 1, 2, 3 \}$, (3) lead time $LT$, (4) flow time $FT$, and (5) the daily output $K$ in units. \emph{Lead time} refers to the total time between customer order generation and fulfilment. The \emph{flow time} refers to the elapsed time from the order beginning of processing by the supplier until fulfilment. Daily records are averaged throughout the day. The $WIP$ and $L_{qi}$ are recorded on an hourly basis, while the remaining parameters and metrics are recorded upon order fulfilment.

\subsubsection{Simulation model validation}

The simulation model is validated using the closed-form model given by \citet{b27} for a particular system configuration. That configuration assumes an infinite number of orders in front of the supplier. In addition, buffer capacity is not allowed at either the manufacturer or the distributor. The calculated rate at which orders leave the system (output rate) for that configuration, using the closed-form model, is compared to the estimated rate from the simulation model.

The simulation model is validated before generating the required data sets to verify the introduced approach. Therefore, a total of $916$ single day replications are used for validation. The calculated output rate from the closed-form model is $10.69$ units per day. The estimated output rate was $10.48 \pm 0.221$ units per day with a $99\%$ confidence level. Moreover, a comparison between the calculated and estimated rates using a Z-test shows no significant difference with a $0.01$ significance level.

\subsubsection{Data sets generation}

Five data sets are generated corresponding to represent five scenarios $S_{i}, i \in \left\{ 0, 1, 2, 3, 4 \right\}$. These scenarios consider both normal and disrupted circumstances. The generated data sets for each scenario represent a multivariate time series that consists of $916$ time records per replication. Each time step includes thirteen parameters (features). These features are (1) interarrival time, (2) supplier processing time, (3) manufacturer processing time, (4) distributor processing time, (5) supplier queue length, (6) manufacturer queue length, (7) distributor queue length, (8) work in process, (9) lead time, (10) flow time, (11) waiting time, (12) processing time, and (13) daily output.

Each disruptive event has a direct impact on some input features in the generated datasets. The surge in demand is represented by a decrease in feature (1) which consequently results in an increase in feature (5), (9), and (11). The second type of disruptive events, capacity loss at any echelon disrupts the whole system and affects features (2–13). For example, considering the capacity loss at the supplier, some of affected features is impacted directly, such as feature (2), and others are impacted indirectly, such as feature (6), because the discontinuity of incoming material flow from the supplier due to the disruptive event.

\subsection{The disruption detection module}

\begin{figure}
	\begin{center}
		\includegraphics[width=\linewidth]{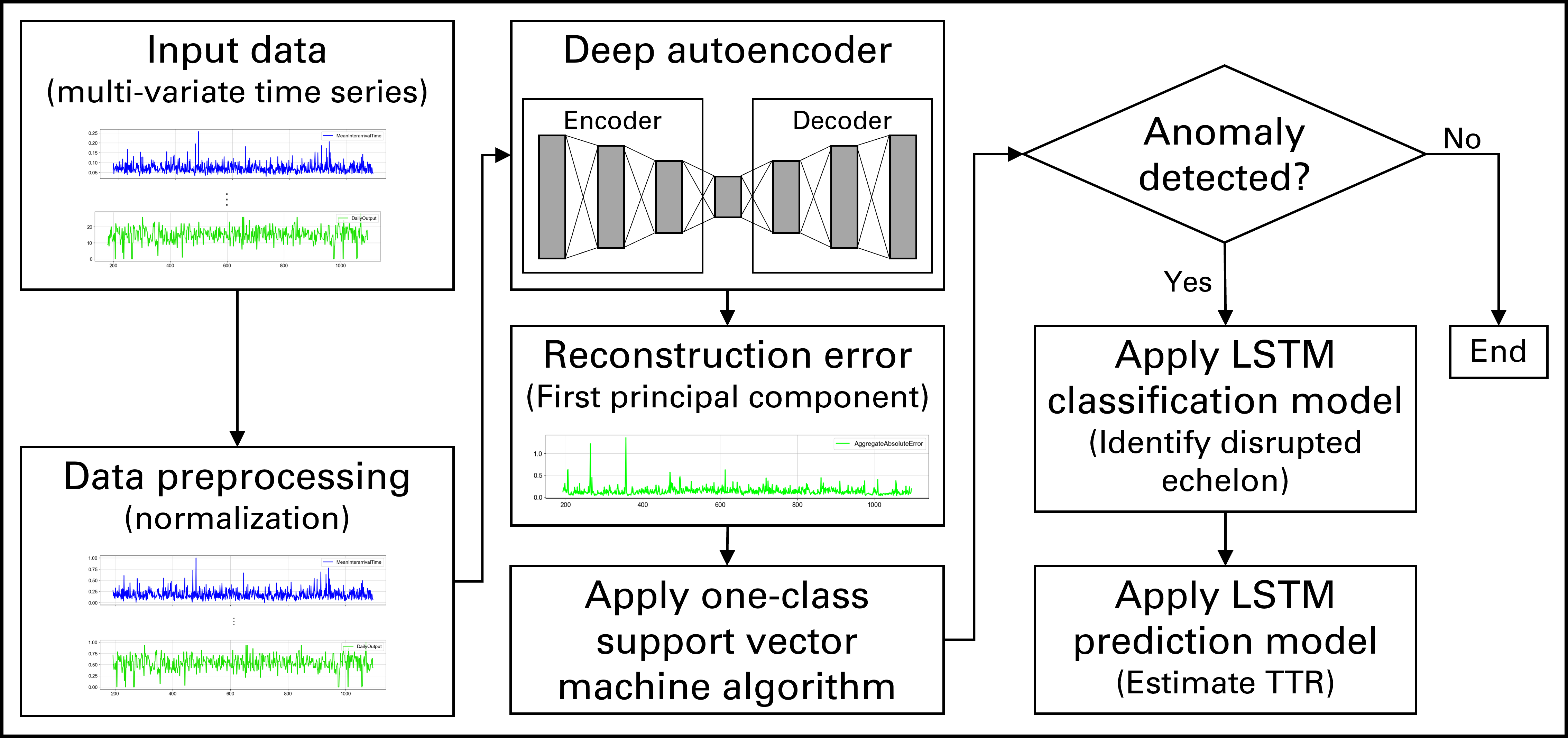}
		\caption{The proposed approach for the disruption detection module in a cognitive digital supply chain twin environment.}
		\label{fig:4}
	\end{center}
\end{figure}

A semi-supervised hybrid deep learning approach is adopted to detect disruptions in the above-mentioned virtual supply chain, as depicted in Figure~\ref{fig:4}. The monitored supply chain parameters and performance metrics produce a multivariate time series with multiple time-dependent variables. Consequently, each variable may depend on other variables besides time dependency, making building an accurate model for disruption detection and TTR prediction a complex task. Therefore, a hybrid deep learning-based approach is adopted to tackle this challenge by using automatic feature extraction and learning of the underlying patterns in the input data.

\subsubsection{Data preprocessing}

TThe input time series data are split into train, validation, and test sets using a split ratio of 60\%, 20\%, 20\%, respectively, for all scenarios. Due to different scales on which input variables are measured, data preprocessing is carried out by normalising the inputs using a min-max scaler, Equation ~\ref{eq1}.
\begin{equation}
	x_{norm}^{i} = \frac{x^{i}-x_{min}^{i}}{x_{max}^{i}-x_{min}^{i}}, i \in \{ 1, 2, ..., k \}
	\label{eq1}
\end{equation}
where $x_{norm}^{i}$ denotes the normalised vector for a time-variate variable. $x_{min}^{i}$ and $x_{max}^{i}$ are the minimum and maximum values of vector $x^{i}$, and $k$ is the number of variables in (length of) the time series. Due to the relatively long time series, a sliding window of size $14$ is applied as a preprocessing step. Afterwards, deep autoencoders and OCSVM algorithm detect disruptions based on the first principal component of the reconstruction error. Moreover, two LSTM neural networks are used to identify the disrupted echelon and predict TTR.

\subsubsection{Disruption detection}

A deep autoencoder of three encoder-decoder pairs is developed to reconstruct the inputs. The hidden and coding layers have a size of $256$, $128$, $64$, and $32$, respectively. The learning rate and batch size are set to $10^{-4}$ and $128$, respectively. The autoencoder is trained for $1000$ epochs using input data considering normal circumstances generated from the scenario $S_{0}$. An epoch refers to a complete pass made by the model on the input data set during training.

In the beginning, the OCSVM algorithm is trained using the first principal component of the obtained absolute error vectors for the test set only under normal circumstances, considering the scenario $S_{0}$. Then, the OCSVM algorithm is tested using the test sets under disrupted circumstances under scenarios $S_{i}$ $\forall i \in \left\{ 1, 2, 3, 4 \right\}$. Model hyperparameters $\nu$ and $\gamma$ are set to $0.025$ and $100$, respectively. The first hyperparameter, $\nu$, controls the sensitivity of the support vectors, while the latter, $\gamma$, controls the boundary shape. High values of $\nu$ lead to a more sensitive model, while high values of $\gamma$ result in an overfit to the training data. At the end of this section, balancing model sensitivity with other performance metrics is discussed.

The OCSVM model results are mapped to a labelled data set for further performance evaluation. The selected performance metrics for the disruption detection model include (1) accuracy, (2) precision, (3) recall, and (4) F1-score. The accuracy, Equation~\ref{eq:12}, describes the overall model performance by calculating the ratio of correctly identified observations to the total observations. The precision, Equation~\ref{eq:13}, determines the ratio of correctly identified normal observations to the total number of normal observations. On the contrary, the recall, Equation~\ref{eq:14}, defines the model sensitivity by realising the ratio of correctly identified normal observations to total observations identified as normal. Finally, the F1-score, Equation~\ref{eq:15}, is a weighted average of precision and recall.
\begin{equation}
	\mathrm{Accuracy = \frac{TP+TN}{TP+FP+FN+TN}}
	\label{eq:12}
\end{equation}
\begin{equation}
	\mathrm{Precision = \frac{TP}{TP+FP}}
	\label{eq:13}
\end{equation}
\begin{equation}
	\mathrm{Recall = \frac{TP}{TP+FN}}
	\label{eq:14}
\end{equation}
\begin{equation}
	\mathrm{F1\textnormal{-}score = 2 \times \frac{Precision \times Recall}{Precision + Recall}}
	\label{eq:15}
\end{equation}
where $\textnormal{TP}$, $\textnormal{FP}$, $\textnormal{FN}$, and $\textnormal{TN}$ are true positive, false-positive, false-negative, and true negative. The true positive refers to the number of correctly identified observations as normal. In contrast, the false-positive represents the number of incorrectly identified observations as normal. False-negative defines the number of abnormal observations that are incorrectly identified as normal. The true negative represents the number of abnormal observations that are correctly identified.

In order to provide the decision-maker with more relevant measures, another two additional performance measures, (1) lag and (2) false-positive percentage, are introduced. The \emph{lag} describes the encountered delay in disruption detection. On the other hand, the ratio of incorrectly classified observations prior to disruption occurrence defines the \emph{false-positive percentage}. These additional performance measures provide a better understanding of the impact of changing model hyperparameters on model performance.

\begin{figure}
	\begin{center}
		\begin{subfigure}{0.49\linewidth}
			\includegraphics[width=\linewidth]{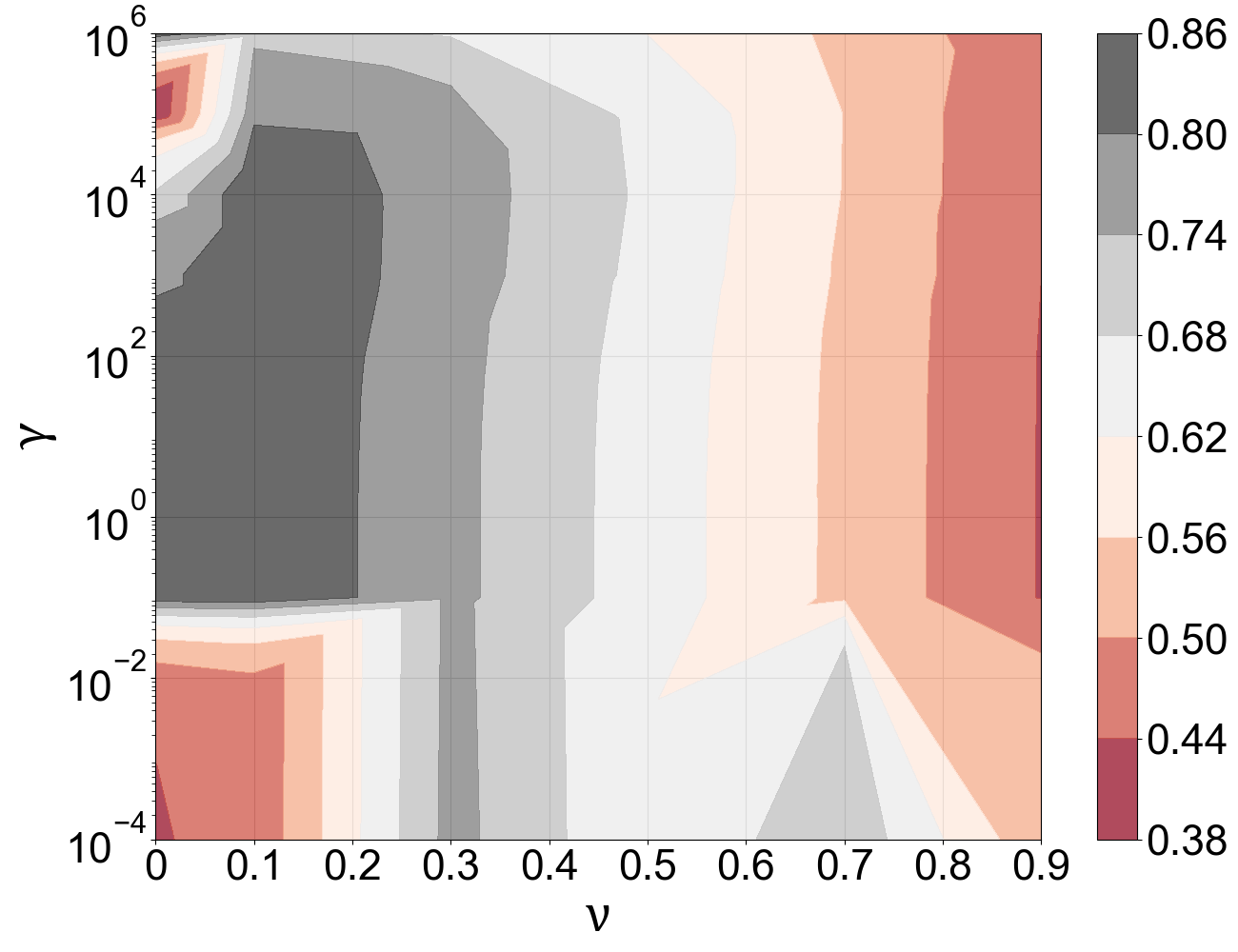}
			\caption{Accuracy.}
			\label{fig:12a}
		\end{subfigure}
		\begin{subfigure}{0.49\linewidth}
			\includegraphics[width=\linewidth]{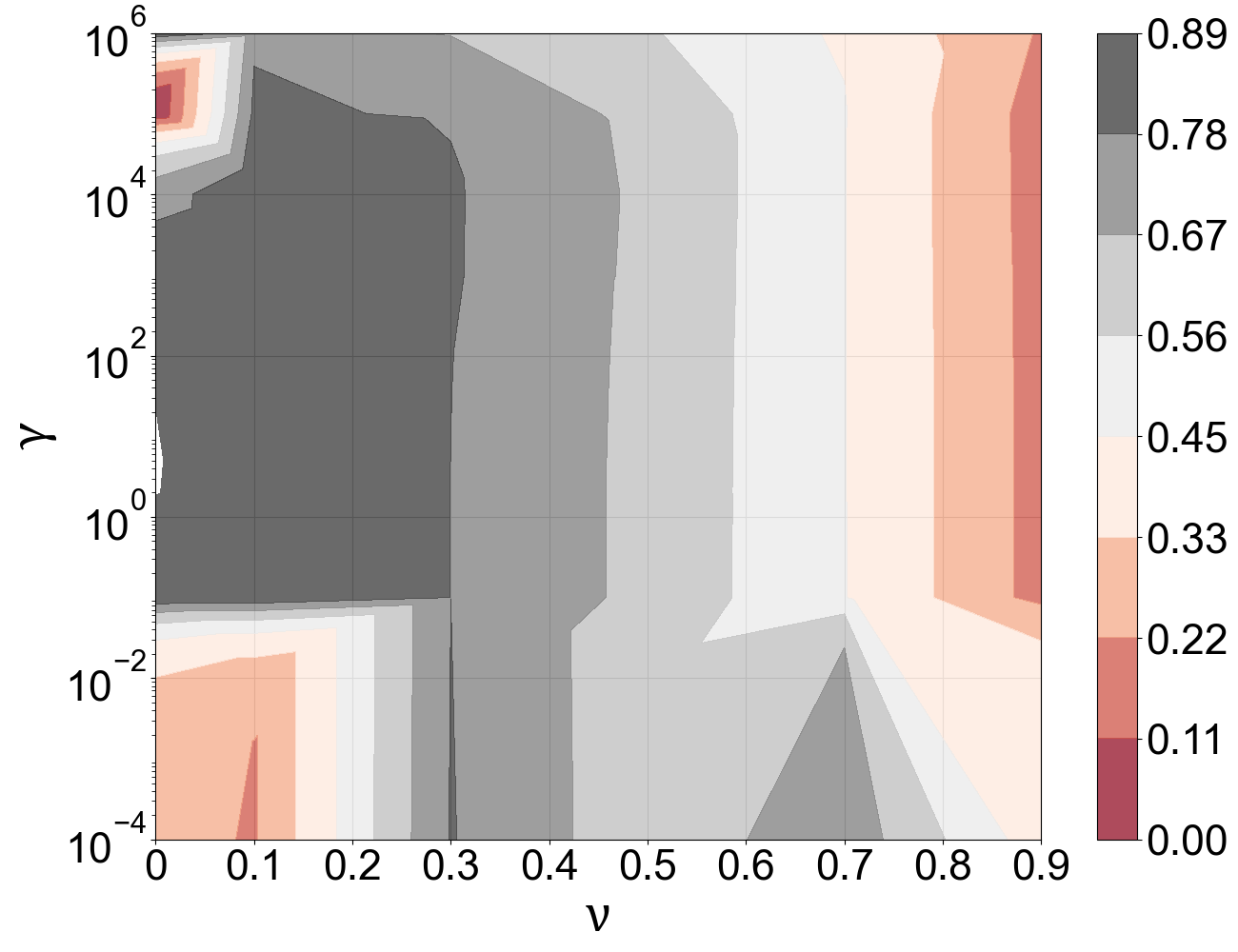}
			\caption{F1-score.}
			\label{fig:12b}
		\end{subfigure}
		\begin{subfigure}{0.49\linewidth}
			\vspace{2mm}
			\includegraphics[width=\linewidth]{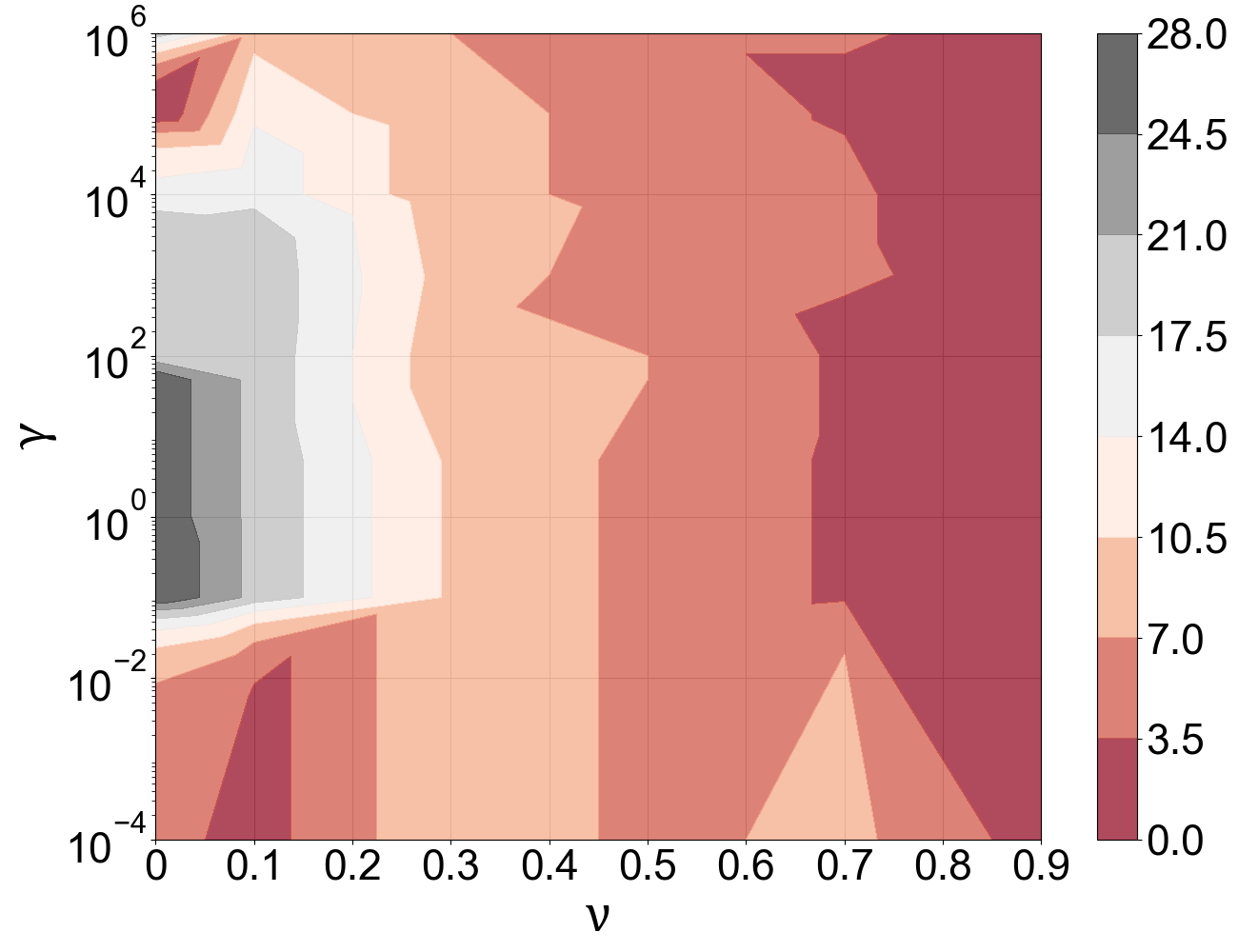}
			\caption{Maximum lag in days.}
			\label{fig:12c}
		\end{subfigure}
		\begin{subfigure}{0.49\linewidth}
			\vspace{2mm}
			\includegraphics[width=\linewidth]{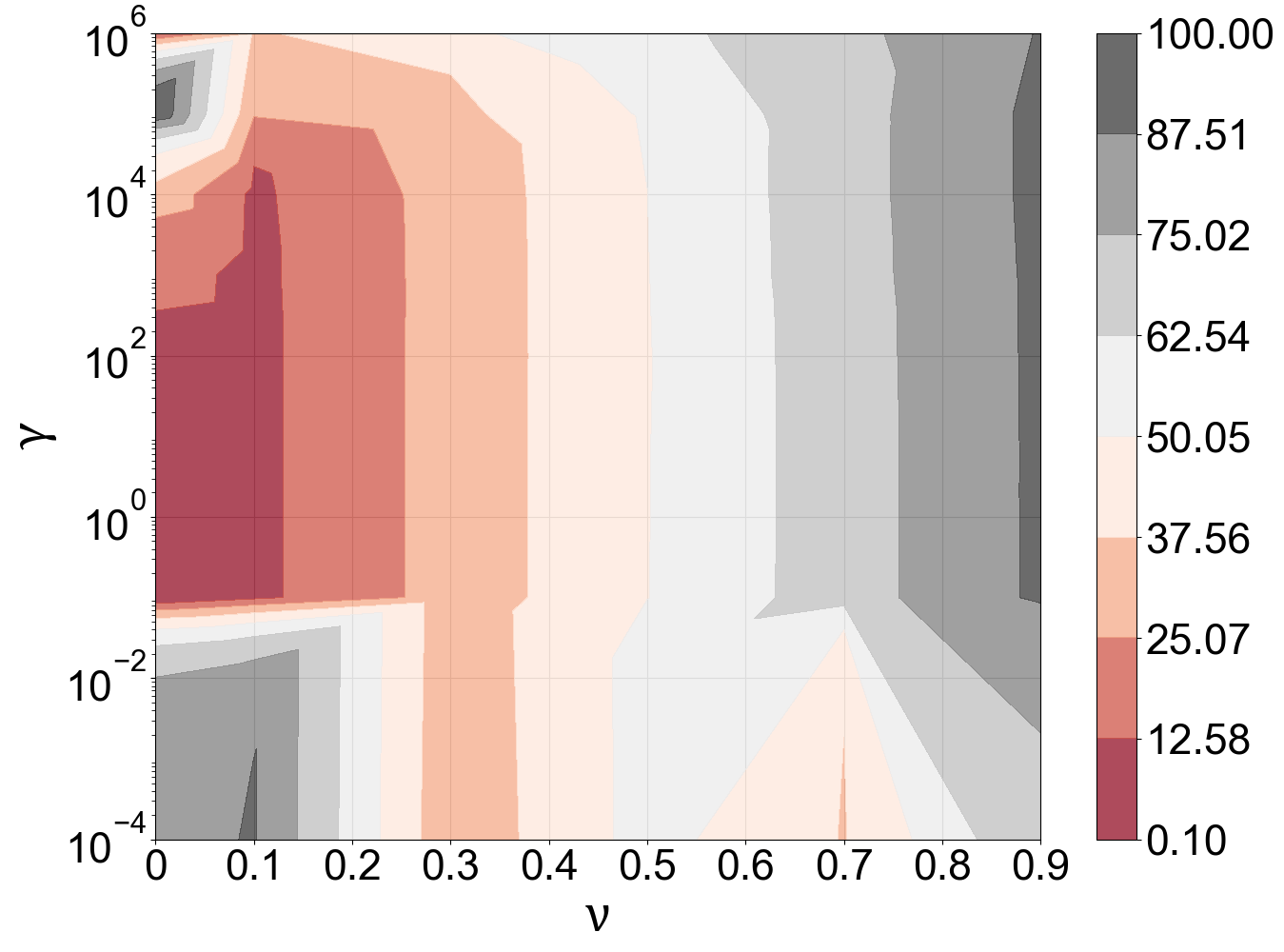}
			\caption{False alarm percentage prior to disruption.}
			\label{fig:12d}
		\end{subfigure}
		\caption{Grid search results for one-class support vector machine hyperparameter selection.}
		\label{fig:12}
	\end{center}
\end{figure}

The OCSVM-based disruption detection model hyperparameters are selected by adopting a grid search approach. The main objective is to find the best performing combination of hyperparameter values based on different performance measures. Figure~\ref{fig:12} summarises the results from the grid search concerning the effect of changing $\nu$ and $\gamma$ values on different performance measures. The x-axis represents $\nu$ on a linear scale, while the y-axis represents $\gamma$ using a log scale. A good model performance can be represented by a combination of high values of accuracy and F1-score in addition to low false alarm percentage. Evidently, better performance is realised at $\nu$ in the range below $0.1$ and relatively moderate values of $\gamma$ between $0.1$ and $100$.

\begin{figure}
	\begin{center}
		\begin{subfigure}{0.49\linewidth}
			\includegraphics[width=\linewidth]{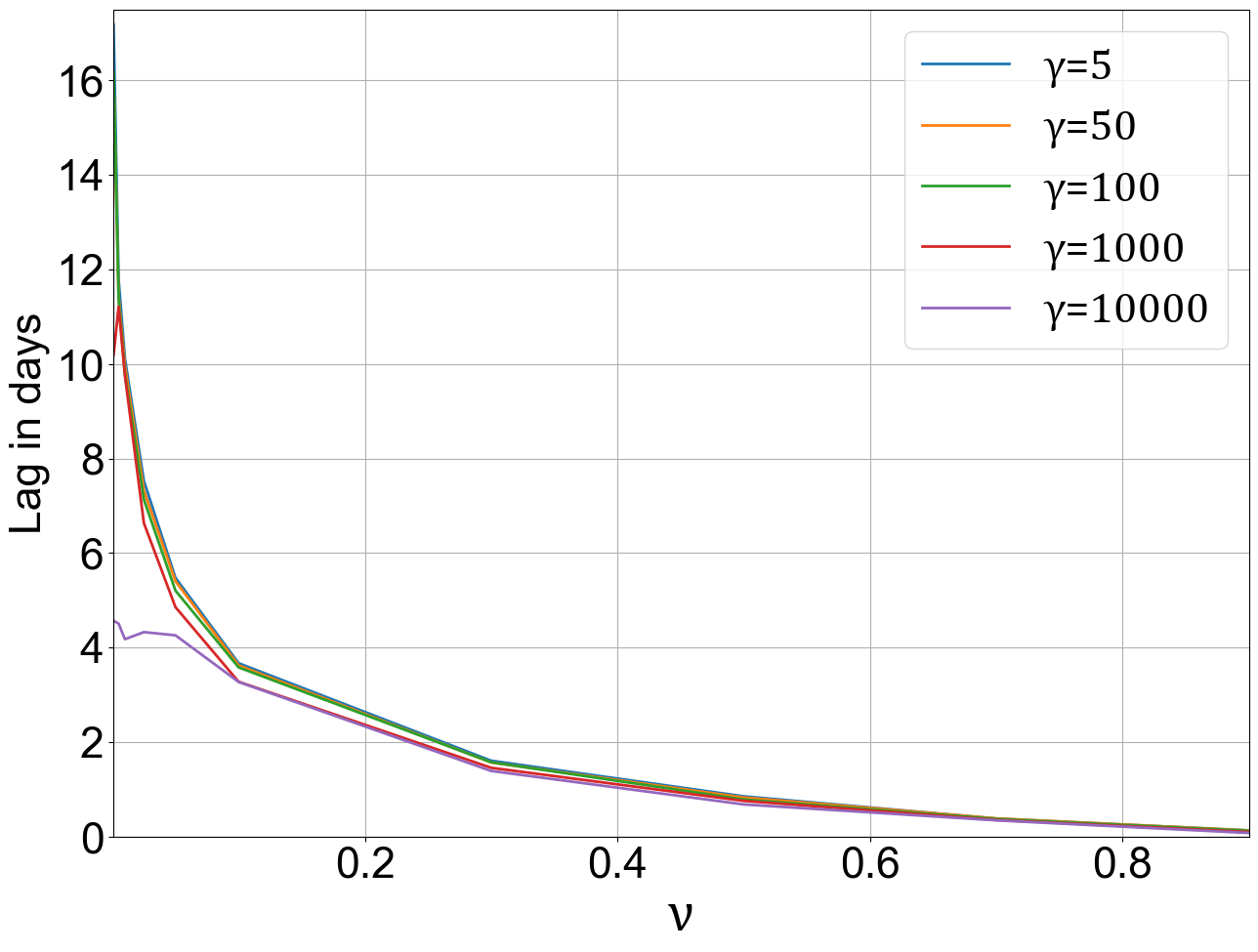}
			\caption{Mean lag at different $\gamma$ values\hfill \break.}
			\label{fig:22a}
		\end{subfigure}
		\begin{subfigure}{0.49\linewidth}
			\hspace{2mm}
			\includegraphics[width=\linewidth]{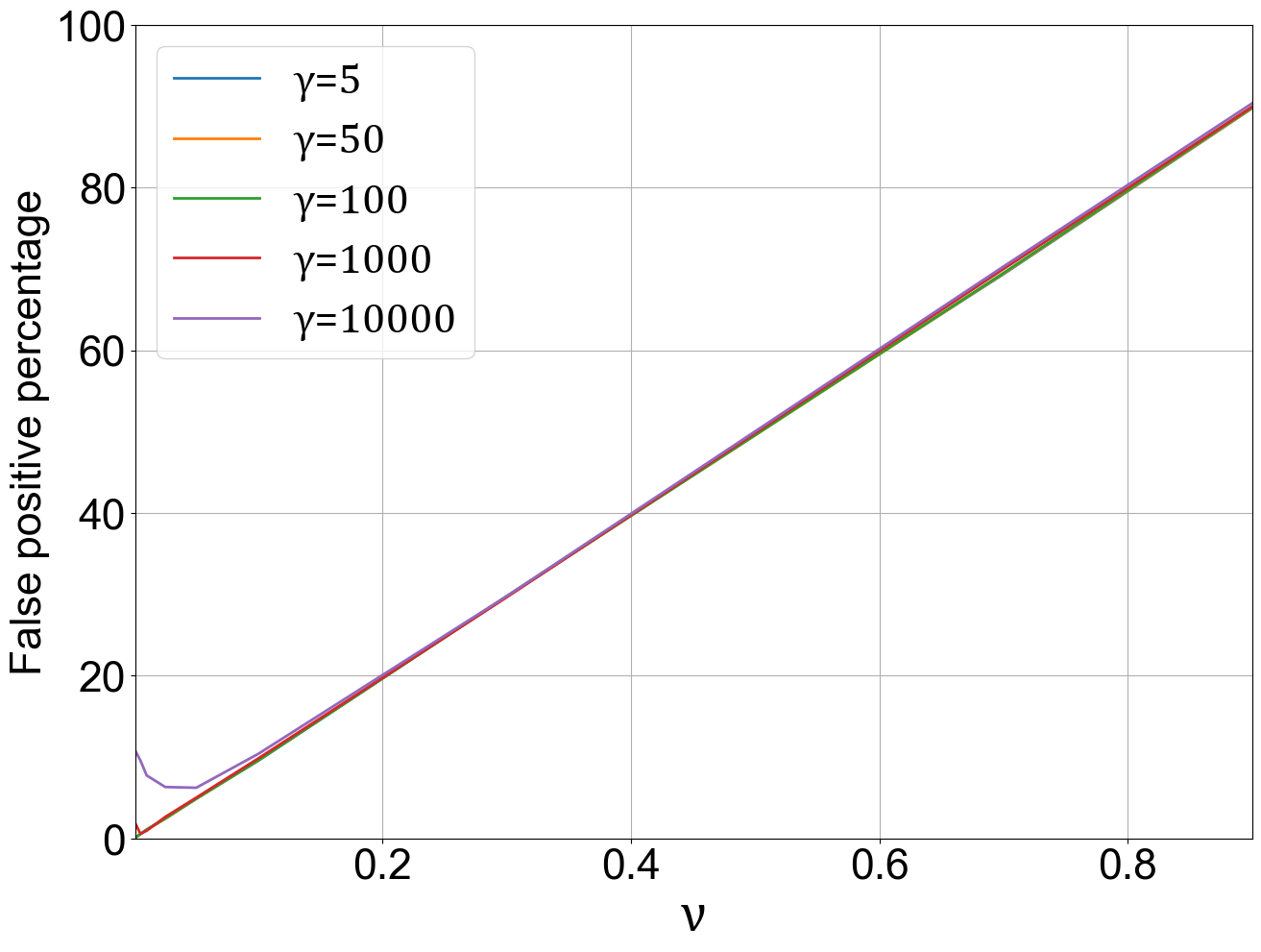}
			\caption{False alarm percentage prior to disruption at different $\gamma$ values.}
			\label{fig:22b}
		\end{subfigure}
		\begin{subfigure}{0.49\linewidth}
			\vspace{2mm}
			\includegraphics[width=\linewidth]{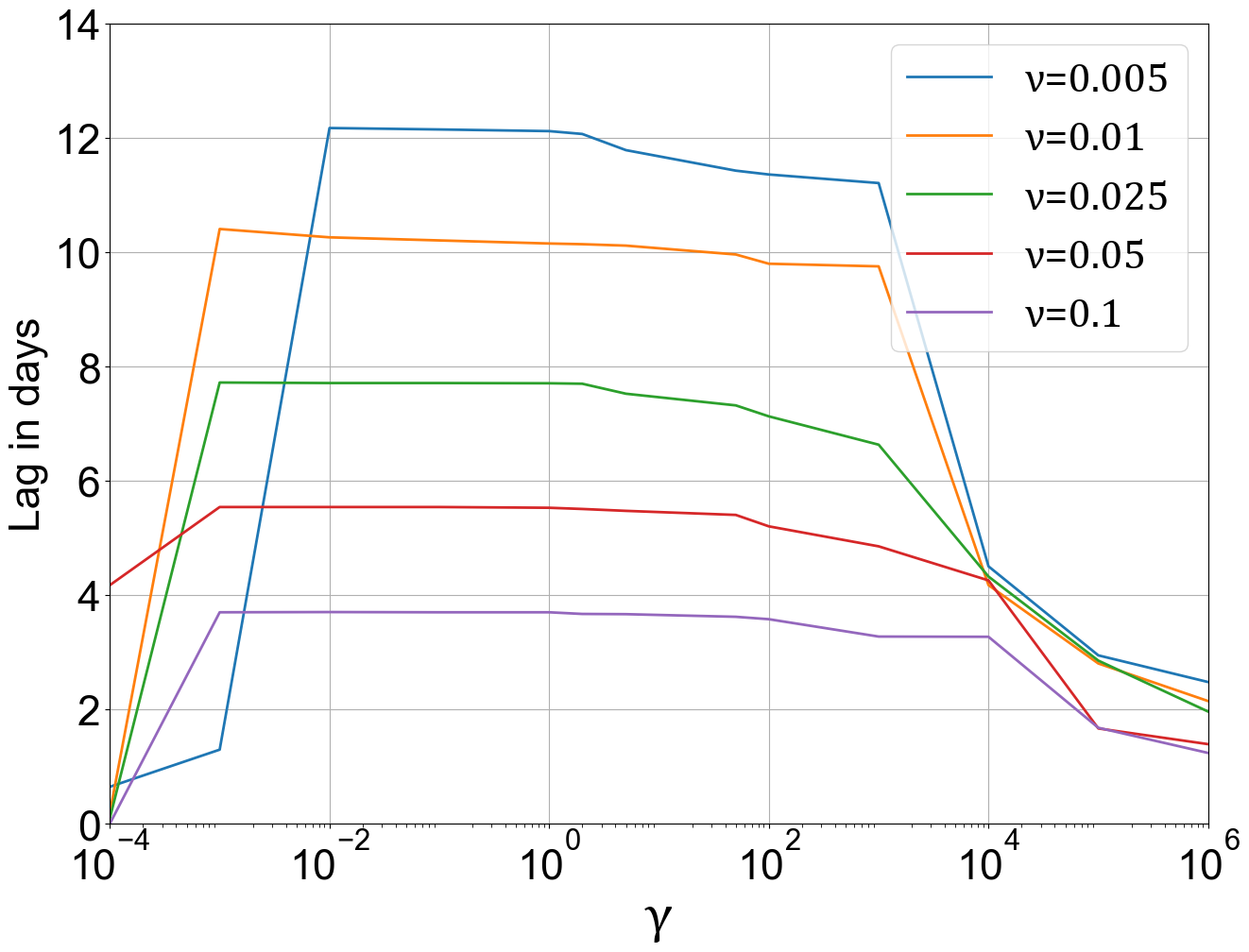}
			\caption{Mean lag at different $\nu$ values\hfill \break.}
			\label{fig:21a}
		\end{subfigure}
		\begin{subfigure}{0.49\linewidth}
			\hspace{2mm}
			\includegraphics[width=\linewidth]{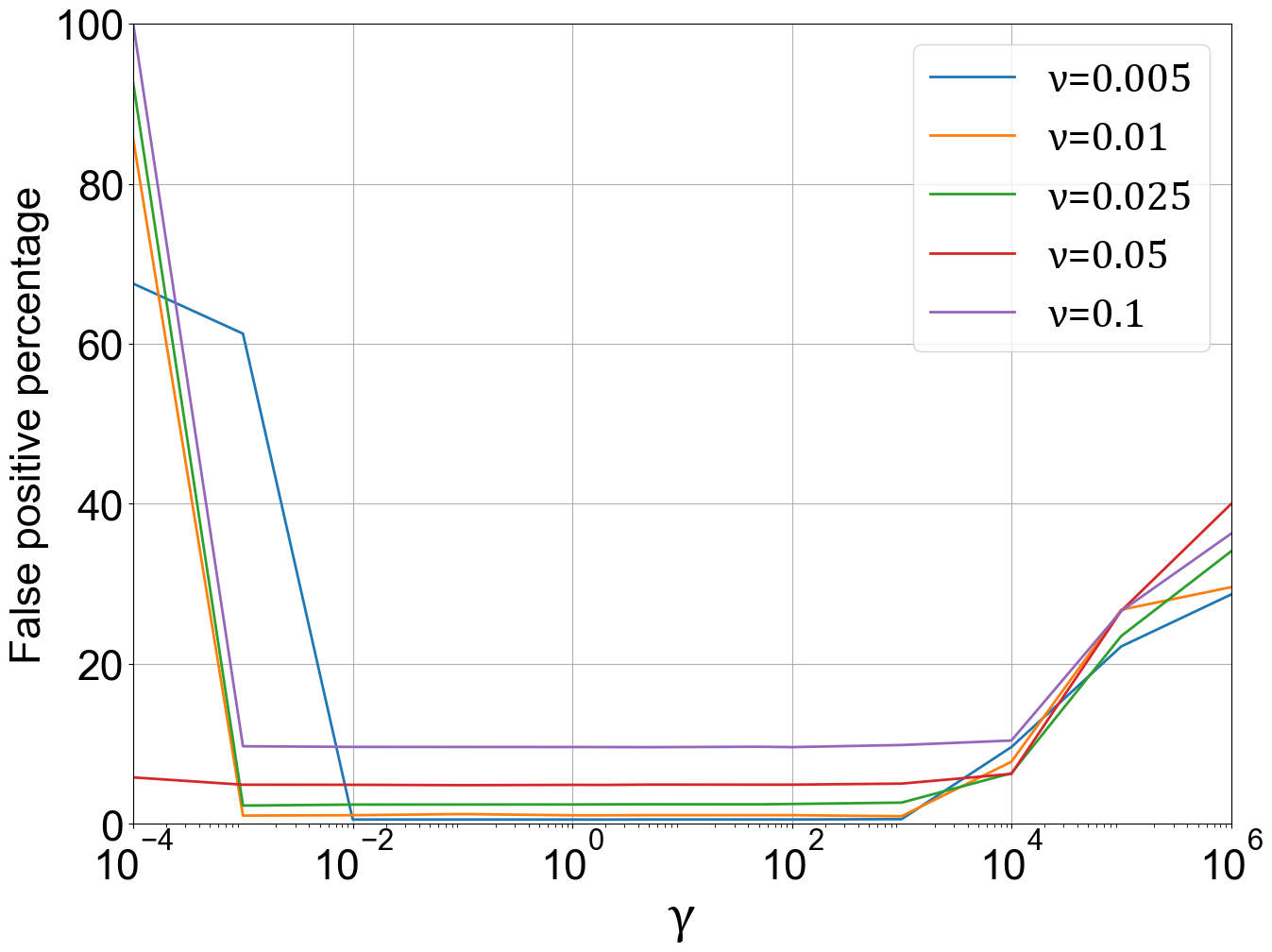}
			\caption{False alarm percentage prior to disruption at different $\nu$ values.}
			\label{fig:21b}
		\end{subfigure}
		\caption{Effect of changing $\nu$ and $\gamma$ values.}
		\label{fig:22}
	\end{center}
\end{figure}

Further analysis is conducted to examine the individual effect of each hyperparameter while the other is fixed on the mean lag and false alarms within the range where good model performance has been observed. Figures~\ref{fig:22a} and~\ref{fig:22b} show the effect of changing $\nu$ while $\gamma$ is fixed at different $\gamma$ values. The x-axis represents $\nu$ while the y-axis represents the performance measure value. As indicated from the shown graphs, $\gamma$ barely affects the performance measures. On the contrary, $\nu$ significantly affects model’s performance at $\nu \le 0.1$.

Figures~\ref{fig:21a} and~\ref{fig:21b} examine the effect of changing $\gamma$ while $\nu$ is fixed at $\nu \le 0.1$. The x-axis represents $\gamma$ on a log scale while the y-axis represents the performance measure value. As per the shown graphs, $\gamma$ does not have a significant effect on the performance measures when compared to $\nu$ at $\gamma \in \left[ 0.01, 1000 \right]$. On the contrary, $\nu$ significantly affects the model’s performance. On the one hand, the increase in $\nu$ results in a significant improvement in the mean lag, but, more false alarms arise. Therefore, the selected values for $\nu$ and $\gamma$ are chosen to achieve as short lags as possible with the fewest false alarms.

\subsubsection{Disrupted echelon identification}

An LSTM neural network classifier is developed to identify the disrupted echelon upon disruption detection. The LSTM classifier is trained in a fully supervised manner. Therefore, the input class sequence is converted from a class vector to a binary class matrix using one-hot encoding. Then, the train and validation sequences are used to train the classifier with a learning rate of $10^{-4}$ and a batch size of $32$ for $20$ epochs. Finally, the classifier is tested using the test set. The LSTM neural network classification model consists of two LSTM layers. Each layer has 16 units and a dropout rate of $0.1$.

\subsubsection{Time-to-recovery prediction}

An LSTM neural network-based model is developed to predict TTR using the incoming signal from the simulation model for different parameters and metrics. Different hyperparameter values are tested, and the best-performing set is chosen to predict the TTR based on the minimum validation loss. These hyperparameters are used to develop four TTR prediction models by considering a single disruption scenario at a time. The four TTR prediction models correspond to the four potential disruption scenarios $S_{i}, i \in \left\{ 1, 2, 3, 4 \right\}$.

Each model has two LSTM layers with 64 LSTM units each. The learning rate is set to $10^{-4}$ and the dropout rate is $0.1$ for each layer. An $l1$ regularisation is applied to the first layer with a regularisation factor of $10^{-3}$. Each model is trained with a batch size of $16$ for twenty epochs. Each model is evaluated based on (1) Mean Absolute Error (MAE), (2) Mean Squared Error (MSE), (3) Root Mean Squared Error (RMSE), and Mean Absolute Percentage Error (MAPE). Each performance measure is given by Equation~\ref{eq:17}, \ref{eq:18}, \ref{eq:19}, and \ref{eq:20}, respectively.
\begin{equation}
	\mathrm{MAE} = \frac{\sum \lvert y - \mathit{\hat{y}} \rvert}{N}
	\label{eq:17}
\end{equation}
\begin{equation}
	\mathrm{MSE} = \frac{\sum \left( y - \mathit{\hat{y}} \right)^{2}}{N}
	\label{eq:18}
\end{equation}
\begin{equation}
	\mathrm{RMSE} = \sqrt{\frac{\sum \left( y - \mathit{\hat{y}} \right)^{2}}{N}}
	\label{eq:19}
\end{equation}
\begin{equation}
	\mathrm{MAPE = \frac{\sum \frac{\lvert y - \mathit{\hat{y}} \rvert}{y}}{N}}
	\label{eq:20}
\end{equation}
where $N$ is the number of TTR observations, while $y$ and $\hat{y}$ represent the actual and predicted TTR vectors.

\section{Results}
\label{sec:results}

The generated data for the virtual supply chain are used to verify the proposed approach. This section evaluates the performance of different modules, mainly the disruption detection module, disrupted echelon identification, and time-to-recovery prediction.

\subsection{Simulation-generated data sets}

After the simulation model is validated, a single data set for each scenario is generated. Then, each data set was labelled and normalised. Finally, each data set was split into train, validation, and test sets. The train and validation sets for scenario $S_{0}$ were used to train the deep autoencoder model. Then, the test set for the $S_{0}$ scenario was used for testing the deep autoencoder model and the OCSVM algorithm. In addition, the test sets for scenarios $S_{i}$ $\forall i \in \left\{ 1, 2, 3, 4 \right\}$ were used for testing the deep autoencoder model, evaluating OCSVM algorithm performance, testing the disrupted echelon classification model, and TTR prediction models. The disrupted echelon classification model and TTR prediction models were trained and validated using the train and validation sets for scenarios $S_{i}$ $\forall i \in \left\{ 1, 2, 3, 4 \right\}$.

\subsection{Disruption detection using deep autoencoders and one-class support vector machine algorithm}

\begin{figure}
	\begin{center}
		\begin{subfigure}{0.49\linewidth}
			\includegraphics[width=\linewidth]{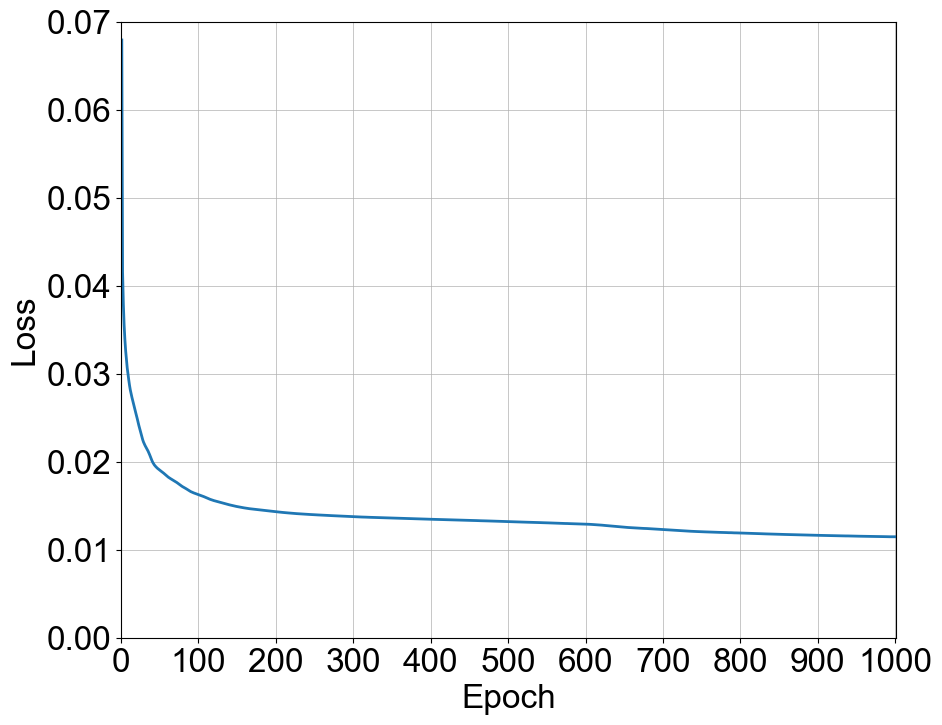}
			\caption{Training loss.}
			\label{fig:9a}
		\end{subfigure}
		\begin{subfigure}{0.49\linewidth}
			\includegraphics[width=\linewidth]{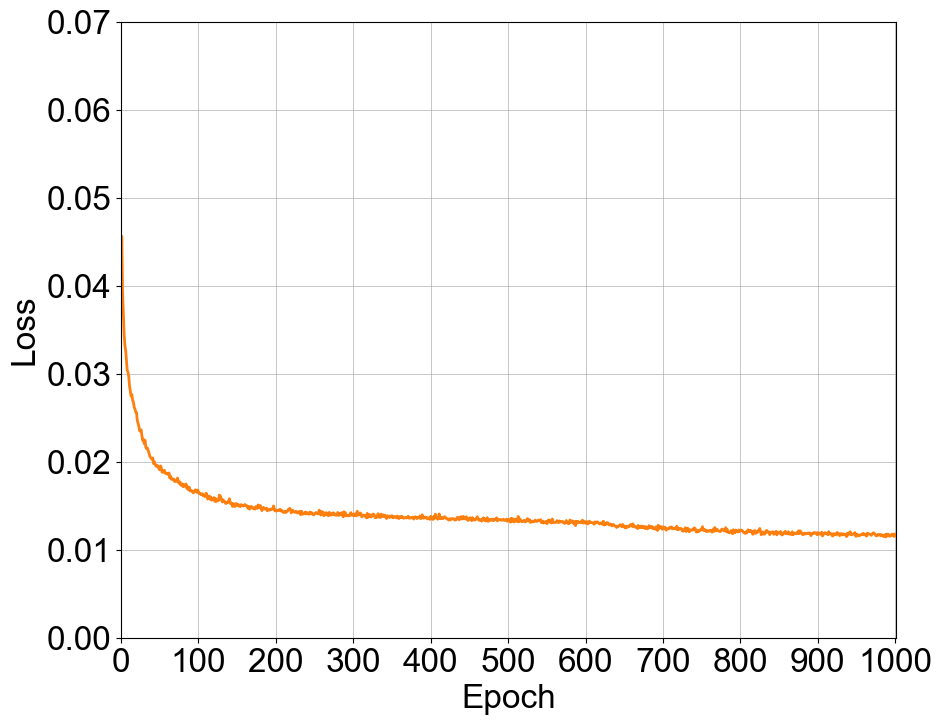}
			\caption{Validation loss.}
			\label{fig:9b}
		\end{subfigure}
		\caption{The learning curve during training the autoencoder.}
		\label{fig:9}
	\end{center}
\end{figure}

The deep autoencoder is trained using sequences of $14 \textnormal{timesteps} \times 13 \textnormal{features}$. These sequences are generated by applying a sliding window of size $14$. Input sequences are converted to a one-dimensional vector due to the inability of the autoencoder to process two-dimensional data as input. The flattened vector has a length of $182$ elements. The MAE Function is used to evaluate the autoencoder model loss. The model loss represents differences between the actual values and the estimations from the model. A learning curve compares the model loss on training and validation data sets. The obtained learning curve demonstrates a slight difference between both data sets, which ensures a good fit of the autoencoder model, Figure~\ref{fig:9}. A significant model loss decrease is noted during the first 100 epochs, followed by a gradual decrease until stability at epoch number $900$.

After training the autoencoder model, it is used to obtain the absolute reconstruction error using the test sets under normal and disrupted circumstances. The absolute reconstruction error $e_{t}^{i}$ for feature $i$ at time $t$ is given by Equation~\ref{eq:11}.
\begin{equation}
	e_{t}^{a} = \sum_{i=1}^{k}\lvert x_{t}^{i} - \mathit{\hat{y}}_{t}^{i} \rvert
	\label{eq:11}
\end{equation}
where $x_{t}^{i}$ and $\hat{x}_{t}^{i}$ are the actual and estimated values of the test set for feature $i$ at time $t$, respectively. A significant difference between the normal and abnormal circumstances was realised due to the low values for the first principal component under normal circumstances. The vast majority of the first principal component values under normal circumstances fall below $-0.2$, which are much lower than those under disruption and recovery, which falls between $-0.5$ and $3.5$.

Then, the OCSVM algorithm is trained using the first principal component vector of the obtained reconstruction error under normal circumstances, which defines the positive class. The first principal component explains $92.39\%$ of the overall variability in the absolute reconstruction error across input features. Finally, the first principal component vector under disrupted circumstances is used for disruption detection using the trained OCSVM disruption detection model. Table~\ref{tab:6} shows the performance evaluation results for the disruption detection model.

\begin{table}
	\begin{center}
		\caption{Performance measures after applying one-class support vector machine algorithm.}
		\label{tab:6}
		\begin{tabularx}{0.8\linewidth}{lcccc}
			Data set & Accuracy & Precision & Recall & F1-score \\
			\hline
			Test-$S_{0}$ & 97.5\% & 100.0\% & 97.5\% & 98.73\% \\
			Test-$S_{i}$ $\forall i \in \left\{ 1, 2, 3, 4 \right\}$ & 87.28\% & 84.25\% & 97.6\% & 90.43\% \\
			\hline
		\end{tabularx}
	\end{center}
\end{table}

There is a considerable difference between the model performance for both data sets. However, the disruption detection model achieved good performance under disrupted circumstances. The high recall value implies that $97.6\%$ of these observations are correctly identified among all normal observations.

Incorrectly classified observations under normal circumstances exist due to the model’s sensitivity to outliers in the train data. The reconstruction error is affected by noise, representing instantaneous disruptions (operational variability). That variability produces extremely low or high values for the principal component of the reconstruction errors under normal circumstances, affecting the OCSVM algorithm performance. Consequently, model sensitivity to such variability is a matter which requires further investigation.

\begin{figure}
	\begin{center}
		\includegraphics[width=0.7\linewidth]{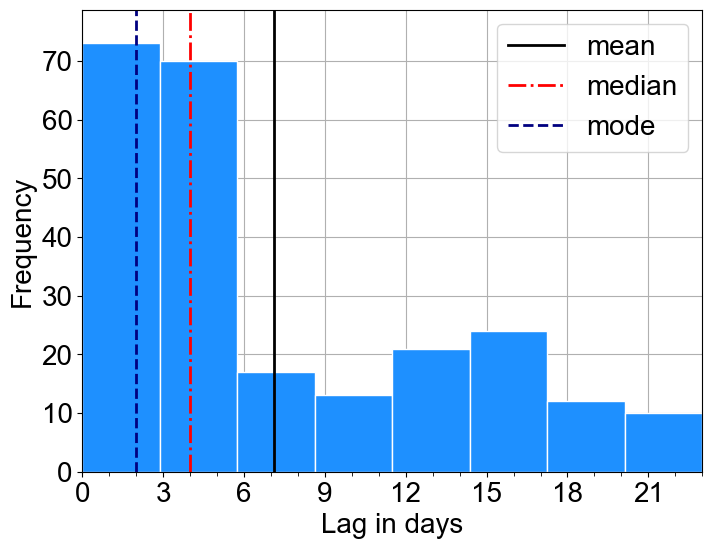}
		\caption{Disruption detection delay distribution.}
		\label{fig:13}
	\end{center}
\end{figure}

The false-positive percentage reflecting the percentage of false alarms prior to disruption is $2.5\%$. The false alarm count is $1530$, roughly corresponding to approximately seven incorrect observations per replication. The average delay in disruption detection (lag) is $7.1$ days. The lag distribution is shown in Figure~\ref{fig:13}. The maximum and median lag values are $23$ and $4$ days, respectively.

Despite the apparent good model performance, the realised lag is a matter of concern depending on the anticipated speed in detecting disruptions. The trade-off between achieving shorter delays and reducing false alarms depends on the model sensitivity, controlled by the hyperparameter $\nu$. Although small hyperparameter values are recommended to achieve few false alarms, the disruption detection model becomes less sensitive to disruptions (anomalies). Thus, a significant increase in maximum lag (delay) is encountered. Large $\nu$ values can achieve an efficient disruption detection model through delay minimization. However, the model becomes too sensitive to , leading to many false alarms and poor performance in terms of accuracy, precision, recall, and F1-score. Therefore, the decision-maker should compromise the combination between the acceptable limits for the performance measures. A suggested solution is to maintain shorter delays. The false alarms can be handled using the proposed LSTM neural network classification model.

\begin{figure}
	\begin{center}
		\begin{subfigure}{0.8\linewidth}%{0.99\linewidth}
			\includegraphics[width=\linewidth]{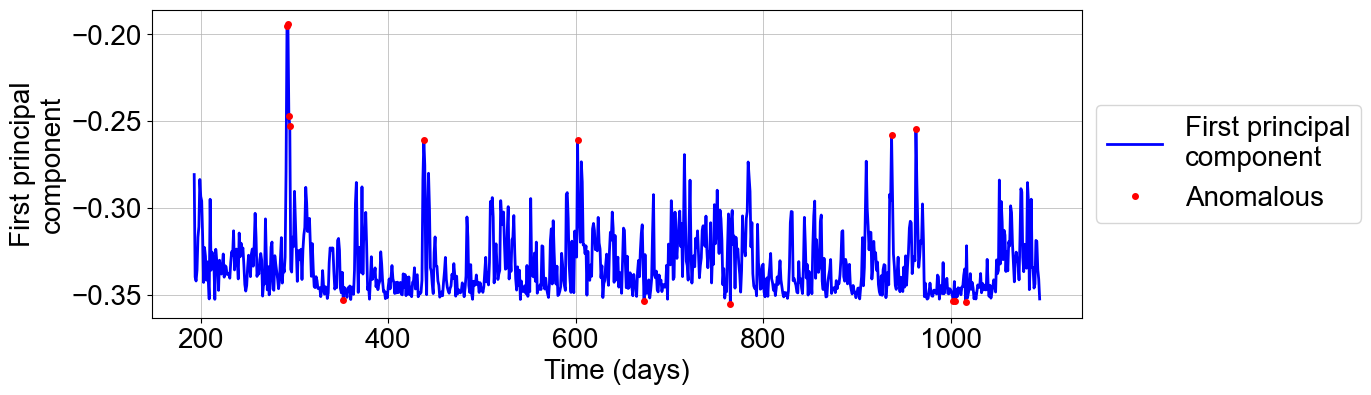}
			\caption{Non-disrupted scenario $S_{0}$.}
			\label{fig:14a}
		\end{subfigure}
		\begin{subfigure}{0.8\linewidth}%{0.49\linewidth}
			\includegraphics[width=\linewidth]{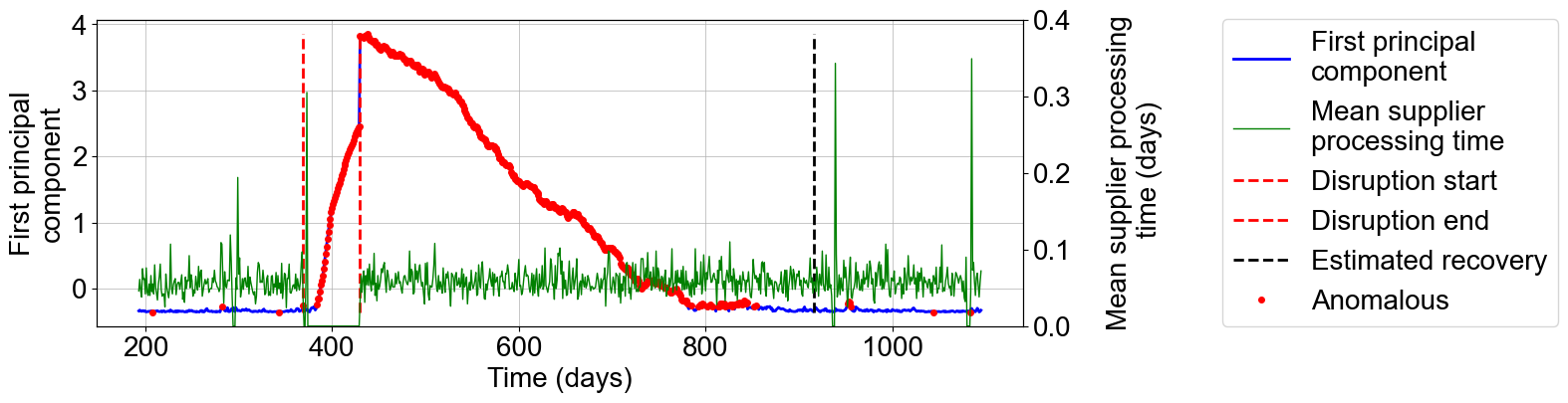}
			\caption{Capacity loss at the supplier scenario $S_{1}$.}
			\label{fig:14c}
		\end{subfigure}
		\begin{subfigure}{0.8\linewidth}%{0.49\linewidth}
			\includegraphics[width=\linewidth]{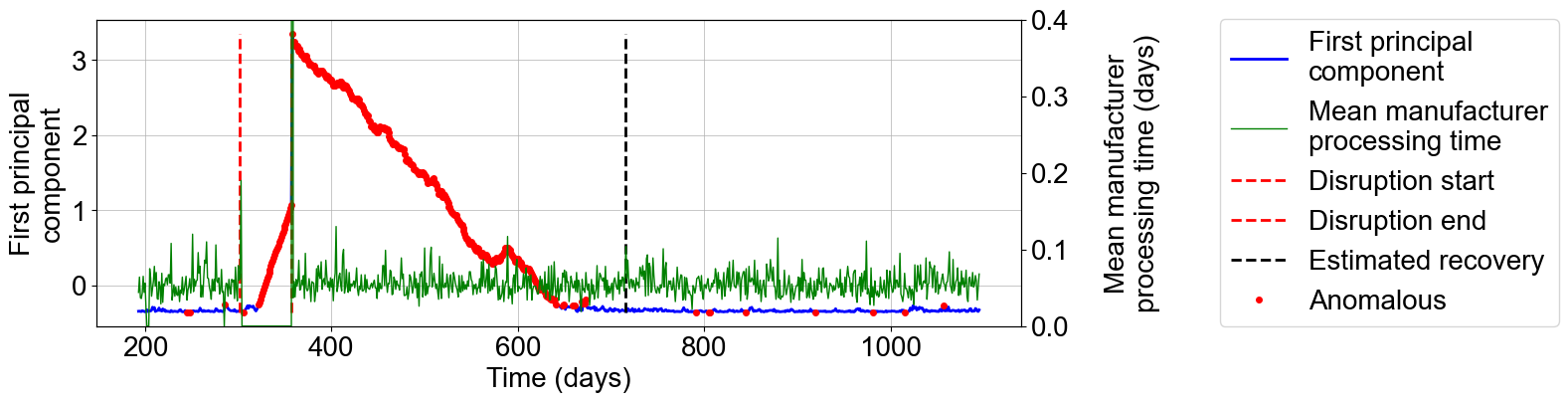}
			\caption{Capacity loss at the manufacturer scenario $S_{3}$}
			\label{fig:14d}
		\end{subfigure}
		\begin{subfigure}{0.8\linewidth}%{0.49\linewidth}
			\includegraphics[width=\linewidth]{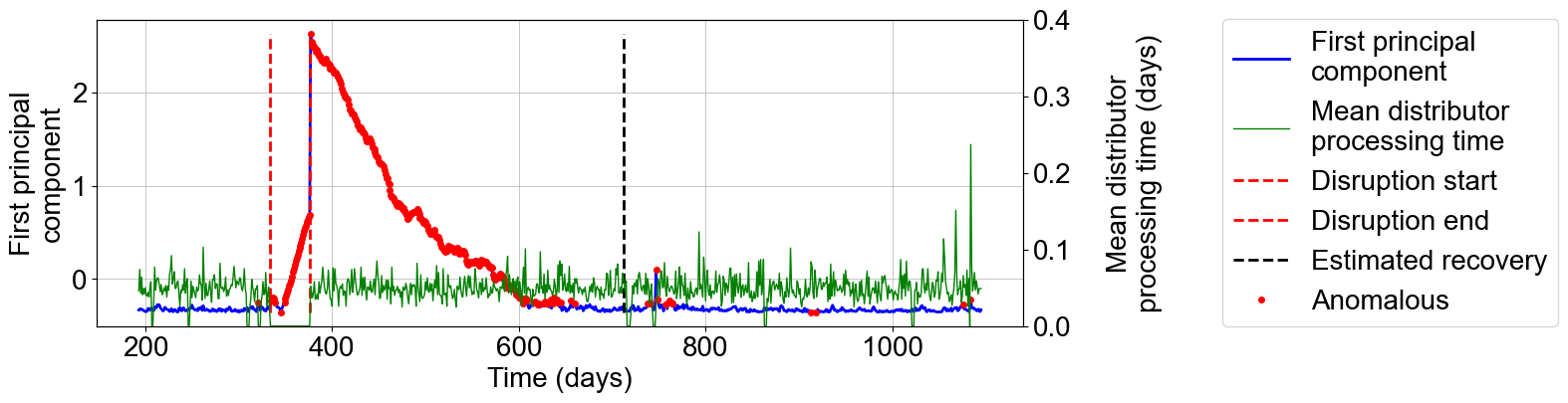}
			\caption{Capacity loss at the distributor scenario $S_{3}$.}
			\label{fig:14e}
		\end{subfigure}
		\begin{subfigure}{0.8\linewidth}%{0.49\linewidth}
			\includegraphics[width=\linewidth]{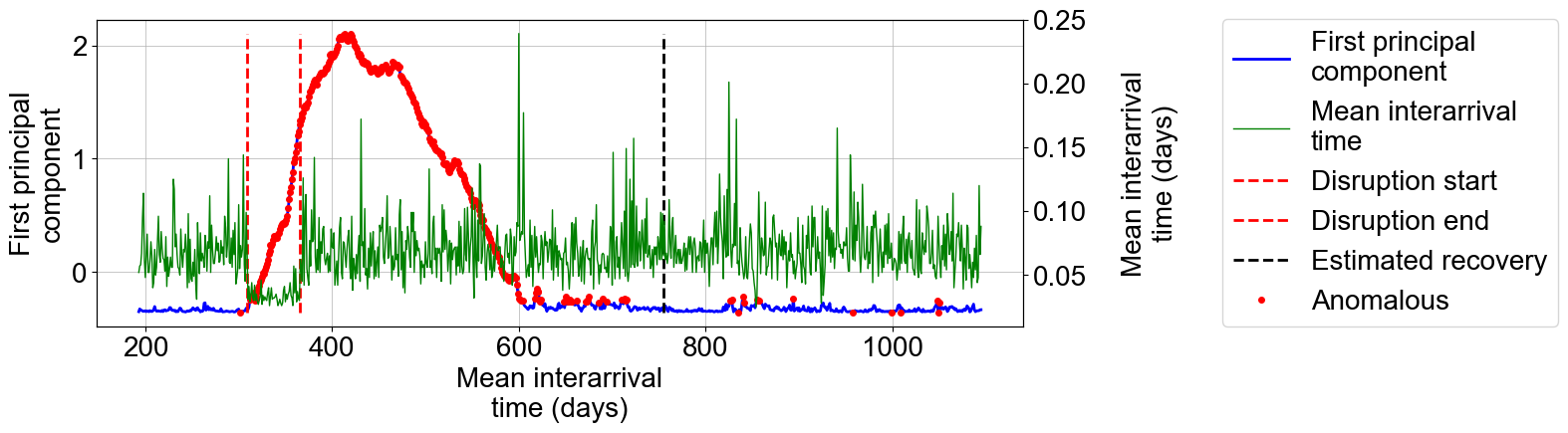}
			\caption{Surge in demand scenario $S_{4}$.}
			\label{fig:14b}
		\end{subfigure}
		\caption{The one-class support vector machine algorithm results.}
		\label{fig:14}
	\end{center}
\end{figure}

The first principal component of the obtained absolute error for a single replication and different scenarios is plotted against time, Figure~\ref{fig:14}. The left y-axis represents the first principal component, while the right y-axis represents the corresponding metric/performance measure for each scenario in days. The first principal component for all disrupted scenarios is notably higher than the scenario under normal circumstances. The red dots refer to the anomalous points. Some points before the estimated recovery are normal points, affecting the model performance measures since the data are labelled based on a predefined threshold.

\subsection{Disrupted echelon identification using long-short term memory neural network model}

\begin{figure}
\begin{center}
\includegraphics[width=0.7\linewidth]{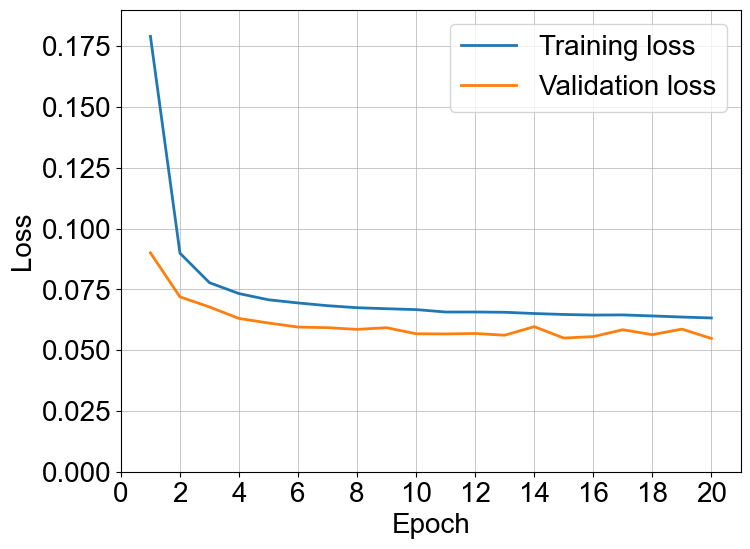}
\caption{The learning curve for long-short term memory classification model.}
\label{fig:15}
\end{center}
\end{figure}

The LSTM model for disrupted echelon identification is trained to learn the multivariate time series pattern. The model is trained using the train and validation data sets for scenarios $S_{i}$ $\forall i \in \left\{ 1, 2, 3, 4 \right\}$. The model should predict the most likely class to which a given sequence belongs. Input data are labelled to consider the disrupted echelon, recovery phase, and normal circumstances during pre-disruption and post-recovery phases. The categorical cross-entropy function $J$, Equation~\ref{eq:16}, is used for model evaluation during training \citep{b46}.
\begin{equation}
	J = -\sum_{k=1}^{N}y_{i, k}.\log \left( p_{i, k} \right)
	\label{eq:16}
\end{equation}
where $N$ is the number of classes, $y_{i, k} \in \left\{ 0, 1 \right\}$ is a binary indicator if class label k is the correct classification for observation $i$, and $p_{i, k} \in \left[ 0, 1 \right]$ is the predicted probability observation $i$ is of class $k$. Lower cross-entropy values indicate better convergence of predicted sample probability towards the actual value. The learning curve shows a significant loss decrease after a few epochs, Figure~\ref{fig:15}.

\begin{figure}
	\begin{center}
		\includegraphics[width=0.9\linewidth]{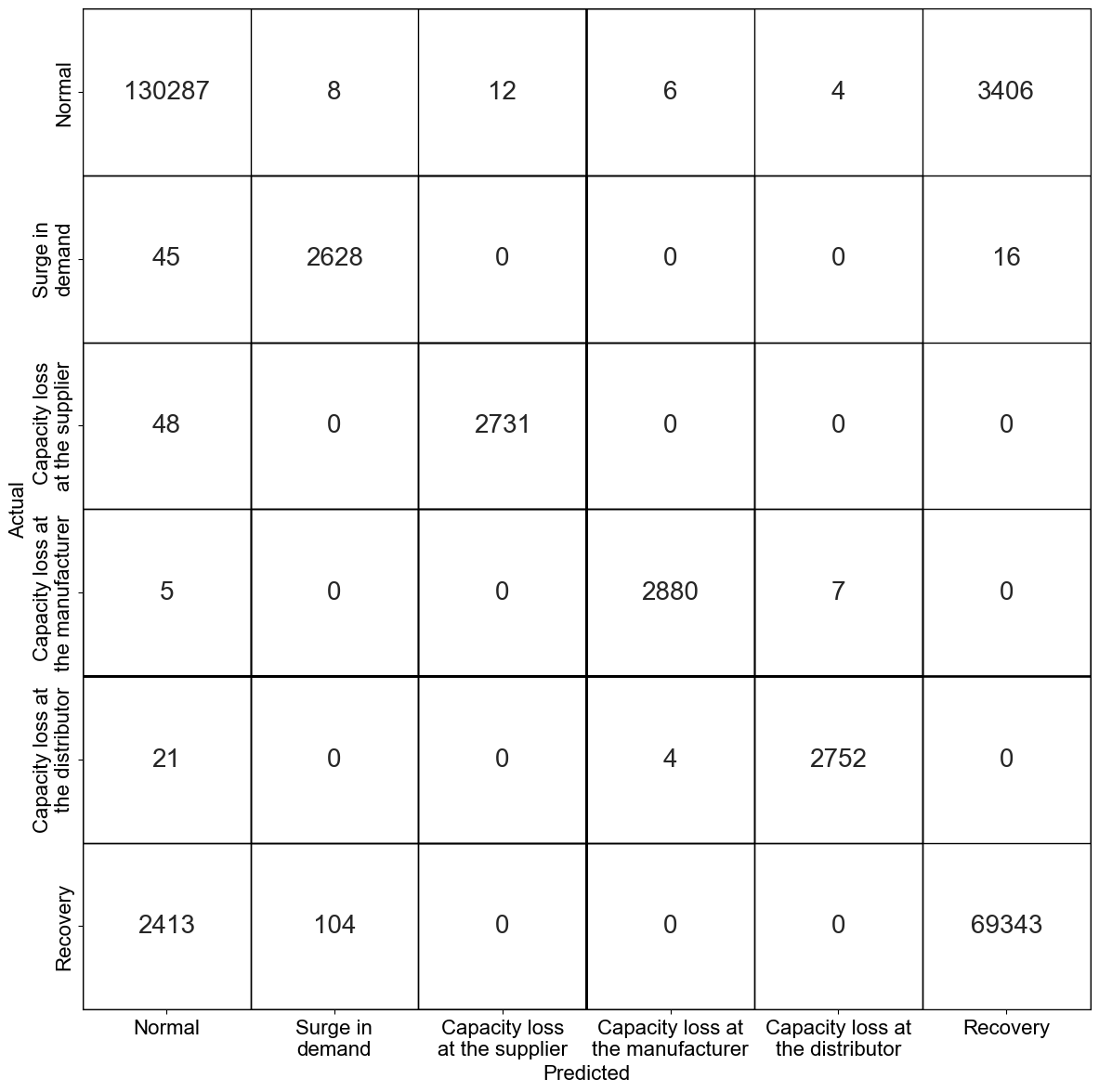}
		\caption{Confusion matrix.}
		\label{fig:20}
	\end{center}
\end{figure}

Once the LSTM neural network model for disrupted echelon identification is trained, it is tested using the test data. The model performance is evaluated using precision, recall, and F1-score. Overall, the model performs well except for identifying the recovery phase, Table~\ref{tab:7}. The precision during recovery is highly affected by the incorrectly classified observations that belong to the normal class, as depicted by the confusion matrix, Figure~\ref{fig:20}. The confusion matrix summarises the LSTM model classification results by showing the count values for each class.

\begin{table}
	\begin{center}
		\caption{Performance measures for long-short term memory classification model.}
		\label{tab:7}
		\begin{tabularx}{0.75\linewidth}{lccc}
			Disruption class & Precision & Recall & F1-score \\
			\hline
			Normal & 98\% & 97\% & 98\% \\
			Surge in demand & 96\% & 98\% & 97\% \\
			Capacity loss at the supplier& 100\% & 98\% & 99\% \\
			Capacity loss at the manufacturer & 100\% & 100\% & 100\% \\
			Capacity loss at the distributor & 100\% & 99\% & 99\% \\
			Recovery & 95\% & 96\% & 96\% \\
			\hline
		\end{tabularx}
	\end{center}
\end{table}

\subsection{Time-to-recovery prediction using long-short term memory neural network models}

\begin{figure}
	\begin{center}
		\begin{subfigure}{0.49\linewidth}
			\includegraphics[width=\linewidth]{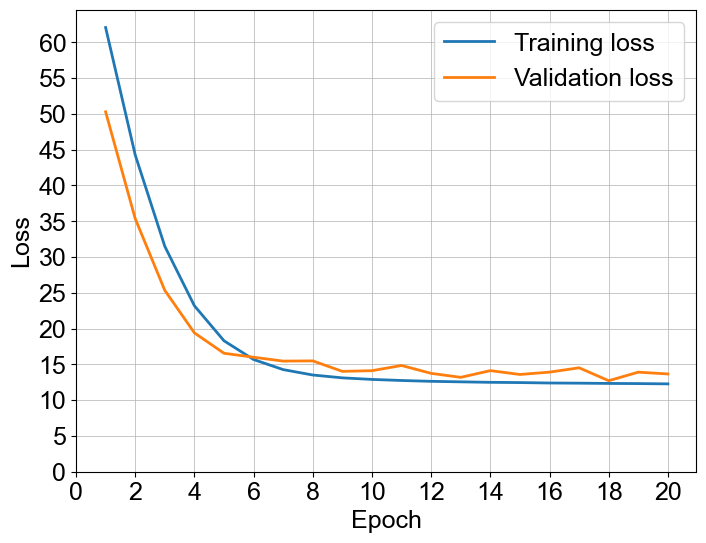}
			\caption{Capacity loss at the supplier scenario $S_{1}$}
			\label{fig:16b}
		\end{subfigure}
		\begin{subfigure}{0.49\linewidth}
			\includegraphics[width=\linewidth]{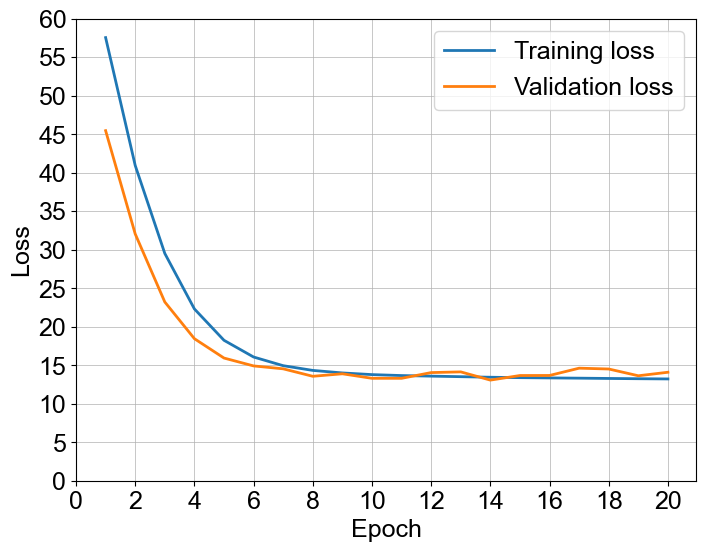}
			\caption{Capacity loss at the manufacturer scenario $S_{2}$}
			\label{fig:16c}
		\end{subfigure}
		\begin{subfigure}{0.49\linewidth}
			\vspace{2mm}
			\includegraphics[width=\linewidth]{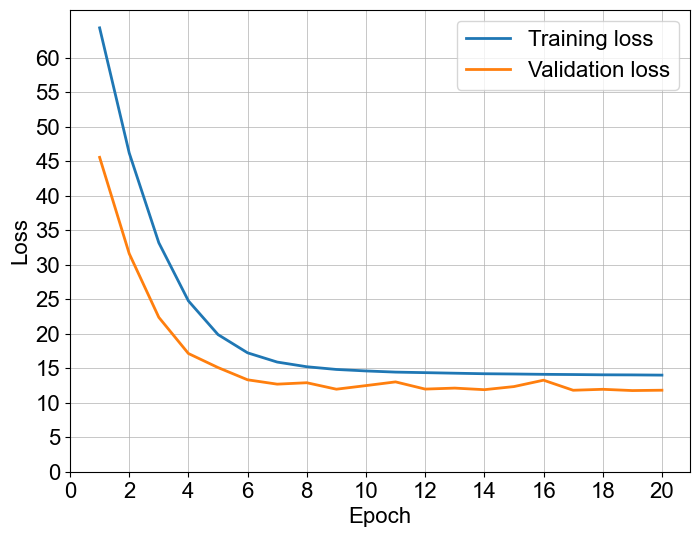}
			\caption{Capacity loss at the distributor scenario $S_{3}$.}
			\label{fig:16d}
		\end{subfigure}
		\begin{subfigure}{0.49\linewidth}
			\includegraphics[width=\linewidth]{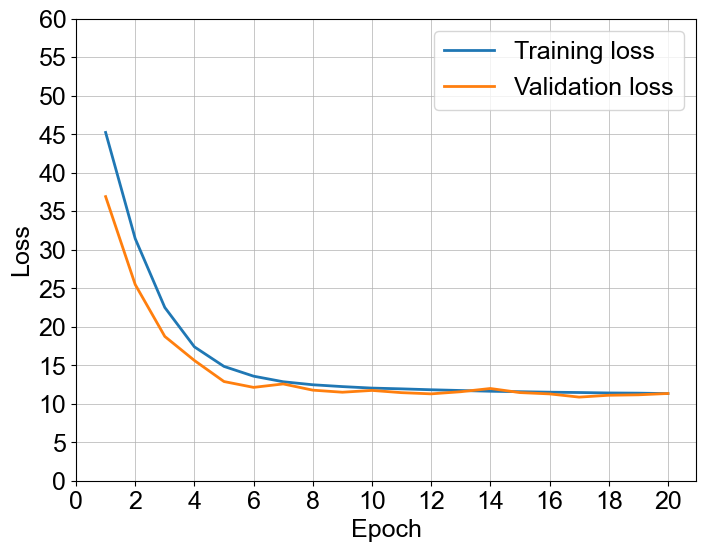}
			\caption{Surge in demand scenario $S_{4}$.}
			\label{fig:16a}
		\end{subfigure}
		\caption{Obtained learning curves for time-to-recovery prediction models.\newline}
		\label{fig:16}
	\end{center}
\end{figure}

The TTR is predicted based on an LSTM neural network prediction model. The model is trained to predict TTR based on multivariate inputs considering a single disruption scenario at a time. Therefore, four prediction models are developed to correspond to each disruption scenario $S_{i}, i \in \left\{ 1, 2, 3, 4 \right\}$. Training and validation data sets are used to train the proposed models. The MAE function monitors the loss for each model. The four models possess a rapid loss decrease after a few epochs, and stability is realised after the eighth epoch, Figure~\ref{fig:16}.

The TTR prediction models are tested using the test sets considering different disruption scenarios $S_{i}, i \in \left\{ 1, 2, 3, 4 \right\}$. It is evident from the performance evaluation results, Table~\ref{tab:8}, that the proposed models perform much better than the results obtained by \citet{b47} for all disruption scenarios. Reducing the number of input features has significantly improved the TTR prediction models performance.

\begin{table}
	\begin{center}
		\caption{Selected error metrics for time-to-recovery prediction models on the test sets.}
		\label{tab:8}
		\begin{tabularx}{0.9\linewidth}{lccccccc}
			\hline
			\multicolumn{1}{c}{\multirow{2}{*}{Scenario}} & \multicolumn{4}{c}{Obtained error measures} & \multicolumn{3}{c}{\citet{b47}} \\
			\cline{2-8}
			\multicolumn{1}{c}{} & MAE & MSE & RMSE & MAPE & \multicolumn{1}{c}{MAE} & \multicolumn{1}{c}{MSE} & \multicolumn{1}{c}{RMSE} \\
			\hline
			$S_{1}$ & 15.32 & 1658.48 & 40.72 & 0.21 & 33.08 & 4142.2 & 64.36 \\
			$S_{2}$ & 17.25 & 1796.42 & 42.38 & 0.235 & 30.52 & 2259.71 & 47.54 \\
			$S_{3}$ & 13.36 & 1193.82 & 34.55 & 0.212 & 43.68 & 5975.85 & 77.3 \\
			$S_{4}$ & 12.8 & 1291.31 & 35.93 & 0.259 & 30.58 & 1867.69 & 43.22 \\
			\hline
		\end{tabularx}
	\end{center}
\end{table}

\begin{figure}
	\begin{center}
		\begin{subfigure}{0.49\linewidth}
			\includegraphics[width=\linewidth]{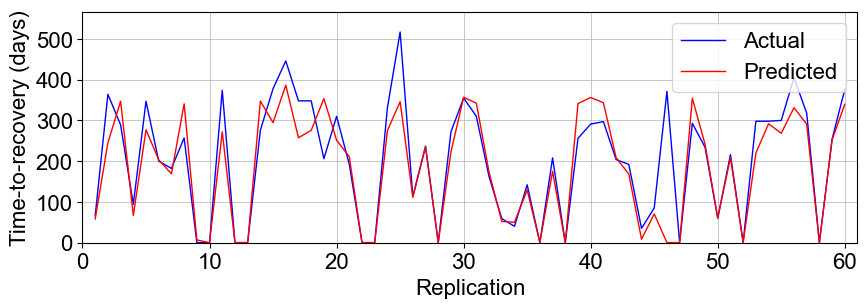}
			\caption{Capacity loss at the supplier scenario $S_{1}$}
			\label{fig:17b}
		\end{subfigure}
		\begin{subfigure}{0.49\linewidth}
			\includegraphics[width=\linewidth]{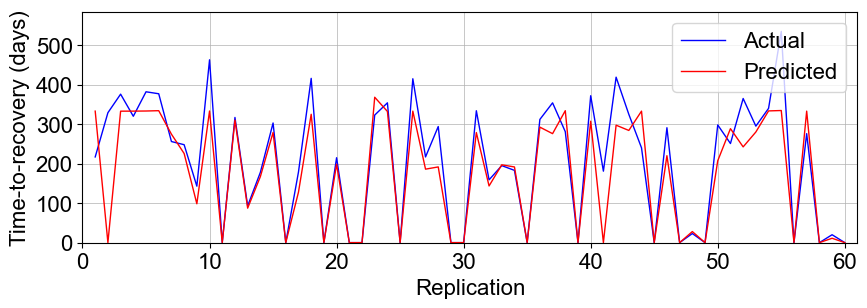}
			\caption{Capacity loss at the manufacturer scenario $S_{2}$}
			\label{fig:17c}
		\end{subfigure}
		\begin{subfigure}{0.49\linewidth}
			\vspace{2mm}
			\includegraphics[width=\linewidth]{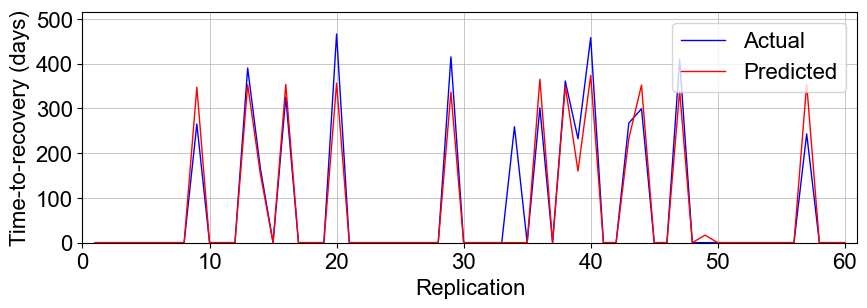}
			\caption{Capacity loss at the distributor scenario $S_{3}$.}
			\label{fig:17d}
		\end{subfigure}
		\begin{subfigure}{0.49\linewidth}
			\includegraphics[width=\linewidth]{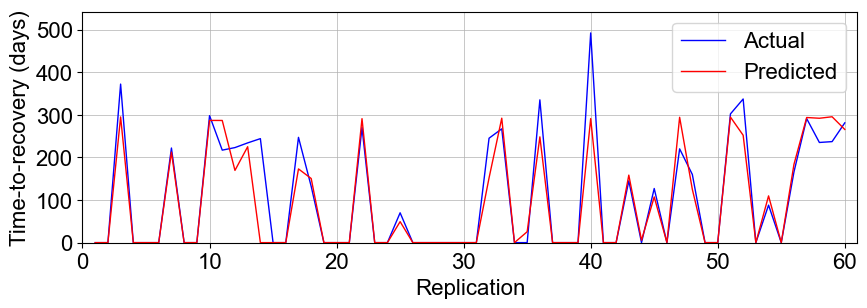}
			\caption{Surge in demand scenario $S_{4}$.}
			\label{fig:17a}
		\end{subfigure}
		\caption{Time-to-recovery predictions versus actual values.}
		\label{fig:17}
	\end{center}
\end{figure}

\begin{figure}
	\begin{center}
		\begin{subfigure}{0.49\linewidth}
			\includegraphics[width=\linewidth]{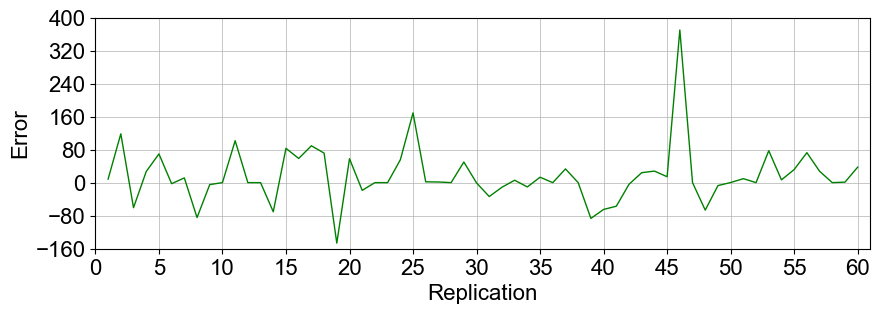}
			\caption{Capacity loss at the supplier scenario $S_{1}$}
			\label{fig:18b}
		\end{subfigure}
		\begin{subfigure}{0.49\linewidth}
			\includegraphics[width=\linewidth]{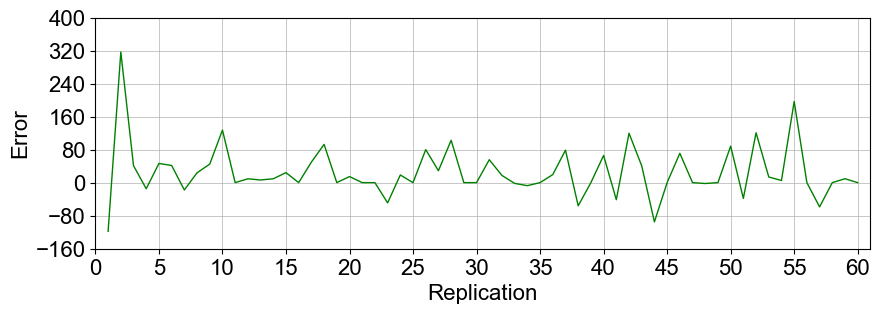}
			\caption{Capacity loss at the manufacturer scenario $S_{2}$}
			\label{fig:18c}
		\end{subfigure}
		\begin{subfigure}{0.49\linewidth}
			\vspace{2mm}
			\includegraphics[width=\linewidth]{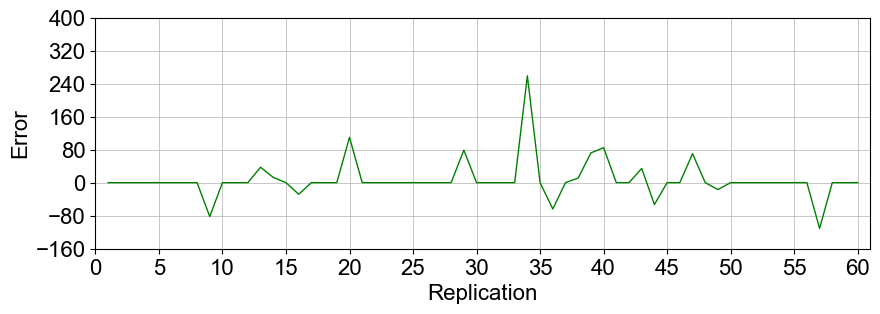}
			\caption{Capacity loss at the distributor scenario $S_{3}$.}
			\label{fig:18d}
		\end{subfigure}
		\begin{subfigure}{0.49\linewidth}
			\includegraphics[width=\linewidth]{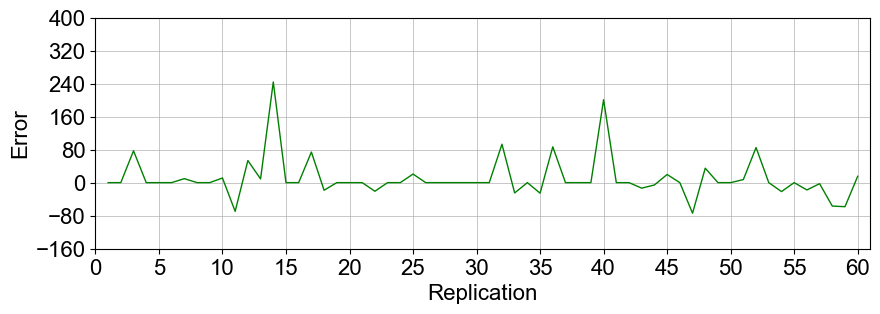}
			\caption{Surge in demand scenario $S_{4}$.}
			\label{fig:18a}
		\end{subfigure}
		\caption{Time-to-recovery prediction errors.}
		\label{fig:18}
	\end{center}
\end{figure}

After the TTR prediction models are tested, the actual and predicted TTR are compared at different replications. The TTR values at a randomly selected time step, $t$, are sketched in Figure~\ref{fig:17}. The predicted TTR values tend to be slightly lower than the actual ones. However, minor variations exist in many cases. The TTR prediction error is obtained by calculating the difference between actual and predicted TTR values. Figure~\ref{fig:18} shows the corresponding prediction error to the data used in Figure~\ref{fig:17}. Significant positive deviations pertain to the early disruption stages.

\begin{figure}
	\begin{center}
		\begin{subfigure}{0.49\linewidth}
			\includegraphics[width=\linewidth]{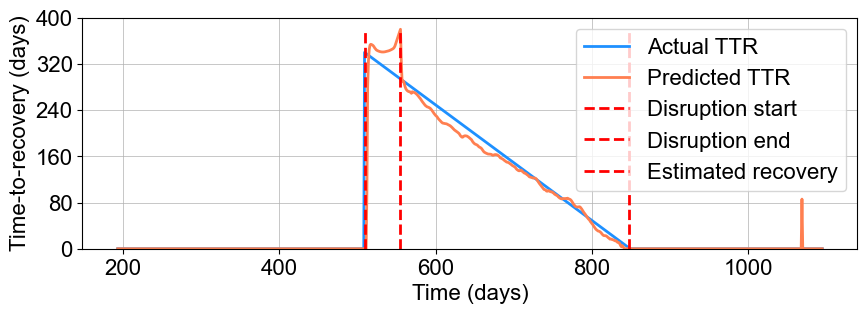}
			\caption{Capacity loss at the supplier scenario $S_{1}$}
			\label{fig:19b}
		\end{subfigure}
		\begin{subfigure}{0.49\linewidth}
			\includegraphics[width=\linewidth]{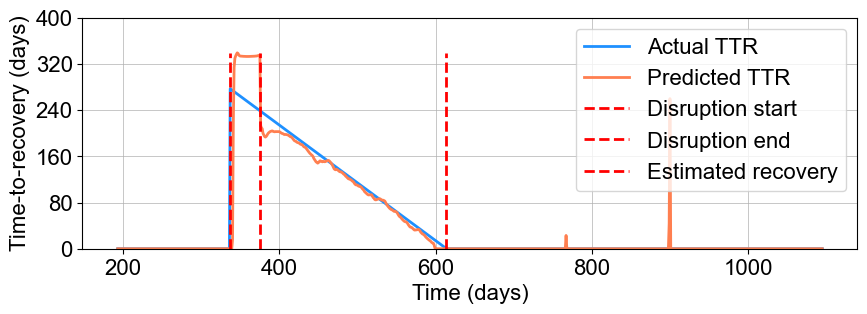}
			\caption{Capacity loss at the manufacturer scenario $S_{2}$}
			\label{fig:19c}
		\end{subfigure}
		\begin{subfigure}{0.49\linewidth}
			\vspace{2mm}
			\includegraphics[width=\linewidth]{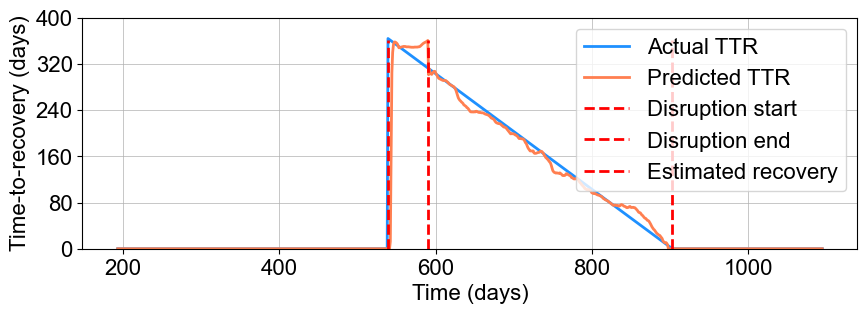}
			\caption{Capacity loss at the distributor scenario $S_{3}$.}
			\label{fig:19d}
		\end{subfigure}
		\begin{subfigure}{0.49\linewidth}
			\includegraphics[width=\linewidth]{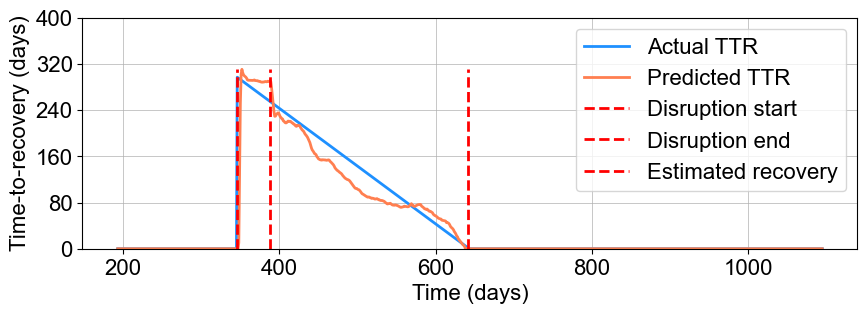}
			\caption{Surge in demand scenario $S_{4}$.}
			\label{fig:19a}
		\end{subfigure}
		\caption{Time-to-recovery prediction evolution along with disruption progression.}
		\label{fig:19}
	\end{center}
\end{figure}

The progression of predicted TTR values is further examined for a single replication considering different disruption scenarios, Figure~\ref{fig:19}. A short delay in TTR prediction is observed at early disruption stages. That delay is followed by a higher TTR prediction than the actual. By the end of the disruption, the predicted TTR values tend to be close to the actual ones.

\section{Managerial implications}
\label{sec:practical implications}

Data-driven driven digital supply chain twins offer better end-to-end supply chain visibility and, consequently, enhanced supply chain resilience. DSCTs can monitor and determine the supply chain state as well as provide useful information and insights for decision-making support. Integrating the proposed models, in this paper, into a cognitive digital supply chain twin helps decision-makers make appropriate decisions based on real-time disruption detection data. Early disruption detection allows for early deployment of recovery policies, minimising negative impact due to disruption, leading to quicker recovery and improved supply chain resilience. In addition, the disrupted echelon identification at early disruption stages allows the decision-makers to find other alternatives that mitigate disruption impact. Furthermore, obtaining the predicted Time-To-Recovery (TTR) at early stages provides an estimate for the duration of contractual agreements, if they exist, when considering different options.

\section{Conclusion}
\label{sec:conclusion}

This paper introduced a new hybrid deep learning-based approach for disruption detection within a data-driven cognitive digital supply chain twin framework. Referring to the first research question “Is there a way to exploit the benefit of cognitive digital twins in the field of supply chain disruption management?” The presented approach mainly contributes to the field of supply chain disruption management by offering better end-to-end supply chain visibility which enhances supply chain resilience through enabling real-time disruption detection, disrupted echelon identification, and time-to-recovery prediction. The developed modules permit the CDSCT to detect disruption occurrence though combining a deep autoencoder neural network with a one-class support vector machine classification algorithm. Then, if a disruption is detected, long-short term memory neural network models identify the disrupted echelon and predict time-to-recovery from the disruption. Referring to the second research question: “How to validate the introduced framework for incorporating cognitive digital twins into supply chain disruption management?” The presented framework is validated under several potential disruption scenarios in a virtual three-echelon supply chain. The disruption scenarios accounted for the surge in demand and unexpected failures at any echelon.

The obtained results indicated a trade-off between disruption detection model sensitivity, encountered delay until disruption detection, and false alarm count. Based on the excellent performance of the proposed model for disrupted echelon identification, that model may be suggested to replace the former approach for disruption detection based on deep autoencoder and one-class support vector machine algorithm. However, the OCSVM algorithm-based anomaly detection model is indispensable because it does not require an extensive definition of all possible disruption scenarios. Developed models for time-to-recovery prediction revealed that predicted time-to-recovery values tend to be lower than the actual ones at early disruption stages. Then, these predictions improve throughout disruption progression with slight variation.

Current research limitations include (1) the difficulty in accurately identifying the transition of the system from a disrupted state to a fully recovered one, (2) considering a single type of disruption at a time, and (3) as a first initiative, the introduced approach has only been tested on simulation-generated data set. Future work directions may include (1) investigating the concurrent occurrence of more than one disruption type, (2) developing a dynamic forecast model to forecast possible supply chain states upon disruption detection, (3) integrating the cognitive digital supply chain twin with an optimization engine to optimize operational decisions to enhance supply chain resilience, (4) examining the performance of other machine learning algorithms, and (5) applying the introduced framework to a real-world case.

\section*{Acknowledgement}

Thanks to Prof. Amin Shoukry (Department of Computer Science Engineering, Egypt-Japan University of Science and Technology, New Borg El-Arab City, Alexandria, Egypt) for his valuable guidance.

\section*{Declarations}

\subsubsection*{Funding}
This work was supported by the Egyptian Ministry of Higher Education (Grant number 10.13039/501100004532) and the Japanese International Cooperation Agency (Grant number 10.13039/501100002385).

\subsubsection*{Conflict of interest/Competing interests}
The authors have no competing interests to declare that are relevant to the content of this article.

\subsubsection*{Availability of data and materials}
The datasets generated during and analysed during the current study are available from the corresponding author on reasonable request.

\subsubsection*{Authors' contributions}
All authors contributed to the study conception and design. The first draft of the manuscript was written by Mahmoud Ashraf and all authors commented on previous versions of the manuscript. All authors read and approved the final manuscript.

\end{document}